\documentclass[11pt]{article}

\usepackage[final]{acl}

\usepackage{times}
\usepackage{latexsym}

\usepackage[T1]{fontenc}

\usepackage[utf8]{inputenc}

\usepackage{microtype}

\usepackage{inconsolata}

\usepackage{graphicx}
\usepackage{float}

\newtheorem{definition}{Definition}[section]

\usepackage{geometry}
\usepackage{booktabs}       
\usepackage{subcaption}
\usepackage{amsmath}
\usepackage{multicol}
\usepackage{algorithm}
\usepackage{algorithmic}
\usepackage{multirow}
\usepackage{makecell}
\usepackage{mdframed}
\usepackage{enumitem}
\usepackage{booktabs}
\usepackage{subcaption}
\usepackage{amsmath}
\usepackage{threeparttable} 

\usepackage{lipsum} 
\usepackage[many]{tcolorbox}
\usepackage{xcolor}
\usepackage{arydshln}

\newtcolorbox{promptbox}[1]{
    enhanced,
    breakable,
    colback=cyan!10!white,
    colframe=cyan!50!black,
    arc=4mm,
    boxrule=1pt,
    fonttitle=\bfseries,
    title=#1, 
    attach boxed title to top left={yshift=-2mm, xshift=3mm},
    boxed title style={
        arc=2mm,
        colback=cyan!50!black,
    }
}

\title{RAFFLES: Reasoning-based Attribution of Faults for LLM Systems}

\author{
  Chenyang Zhu \quad Spencer Hong \quad Jingyu Wu \quad Kushal Chawla \quad Charlotte Tang \\
  \textbf{Youbing Yin \quad Nathan Wolfe \quad Erin Babinsky \quad Daben Liu }\\
  Capital One \\
  \texttt{\{chenyang.zhu, spencer.hong, jingyu.wu, kushal.chawla, yuhui.tang} \\
  \texttt{youbing.yin, nathan.wolfe, erin.babinsky, daben.liu\}@capitalone.com}
}

\begin{document}
\maketitle
\begin{abstract}

The advent of complex, interconnected long-horizon LLM systems has made it incredibly tricky to identify where and when these systems break down. Evaluation capabilities that currently exist today 
are limited in that they often focus on simple metrics, end-to-end outcomes, and are dependent on the perspectives of humans. In order to match the increasing complexity of these many component systems, evaluation frameworks must also be able to reason, probe, iterate, and understand the nuanced logic passing through these systems. In this paper, we present RAFFLES, an offline evaluation architecture that incorporates iterative reasoning. Specifically, RAFFLES operates as an iterative, multi-component pipeline, using a central Judge to systematically identify faults and a set of specialized Evaluators to assess the quality of the candidate faults as well as the rationale of the Judge.  
We evaluated RAFFLES with several benchmarks: the Who\&When datasets to identify step-level faults in multi-agent systems 
and the ReasonEval datasets to diagnose step-level mathematical reasoning errors.  RAFFLES outperforms strong baselines, achieving an accuracy of over 20\% and 50\% on the Who\&When Hand-Crafted and Algorithmically-Generated datasets, and over 80\% on the ReasonEval datasets. These results demonstrate a key step towards introducing automated fault detection for autonomous systems over labor-intensive manual review.
\end{abstract}

\section{Introduction}
\begin{figure*}
  \centering
 \includegraphics[width=\textwidth]{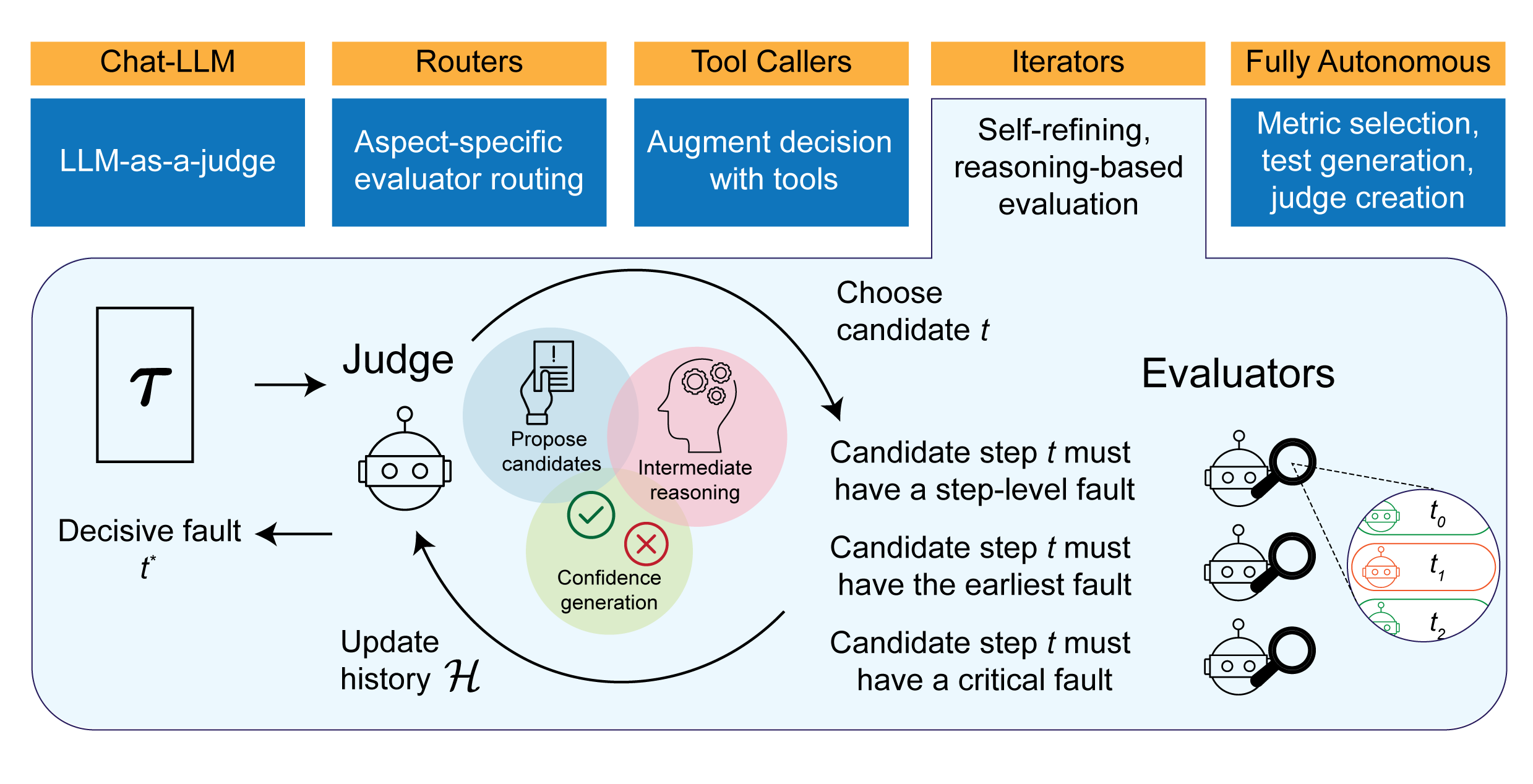}
  \caption{\textbf{Our proposed RAFFLES framework for multi-turn agentic evaluation by reasoning-based fault attribution.} 
  RAFFLES leverages specialized Evaluators designed to assess candidate faults based on the criteria of a decisive fault. Each Evaluator takes in the full log $\tau$ and intermediate reasoning, which are passed to subsequent iterations until a decisive fault is determined. 
  }
  \label{fig:schematic}
\end{figure*}

As large language models (LLMs) evolve into complex, multi-component systems, a critical gap has emerged between what they can do and how we can effectively evaluate them. The rise of language-conditioned agentic architectures such as ReAct \cite{yao2023react}, Toolformer \cite{schick2023toolformer}, and Reflexion \cite{shinn2023reflexion} allows systems to plan, reason, and act over long horizons. However, these new architectures also introduce novel failure modes that current evaluation strategies are not equipped to detect. Most current strategies are confined to isolated metrics that focus on overall outcomes, and they struggle with longer contexts \cite{epperson2025interactive, liu2023lost, xu2025does}. This leaves manual "detective work" as the only way to identify root cause errors. In long-horizon agentic systems, a single, subtle error can quickly cascade, creating a ripple effect of unexpected behaviors. Finding these root cause errors is a true "needle in the haystack" problem. To date, when such step-level evaluation is done manually, it takes on the order of many minutes to tens of minutes \textit{per data instance} \cite{epperson2025interactive, zhang2025agent, deshpande2025trail}. Such manual debugging is too costly and can become a bottleneck. Automation can significantly reduce the manual diagnostic burden, provided these post-hoc implementations transition from measuring end-to-end performance to better understanding \textit{where} failures originate and \textit{when} they form.

Automatic failure attribution within agentic systems is an emerging area of focus for the evaluation community \cite{zhang2025agent, zhou2025shielda, deshpande2025trail, cemri2025multi, guan2025monitoring}. However, preliminary results have demonstrated limited success due to the challenges imposed by detecting faults in complex systems. For example, the LLM-as-a-judge has become a cornerstone of automated evaluation due to its flexibility in approximating human evaluators \cite{zheng2023judging, fu2023gptscore}, and yet single-pass LLM evaluators struggle to detect faults within the long trajectories of agentic systems \cite{zhang2025agent, deshpande2025trail}.  Effective evaluation by which one can pinpoint faulty planning, logic, tool calls, calculations, code, and more, within multi-component systems \textit{requires} its own set of capabilities (e.g., reasoning, planning, tool calling, iterating). One recently proposed framework groups LLM systems into five levels of increasing complexity: Chat-LLM, Routers, Tool-Callers, Iterators, and Fully Autonomous systems \cite{tutorial}. We believe that evaluation systems must likewise evolve in complexity in a parallel manner: from single judgment Chat-LLMs, to aspect-specific evaluator routing, evaluations with tools, iterators that loop and refine, and eventually, fully autonomous evaluators. We present this evolution of evaluators in Figure \ref{fig:schematic}. 

The need to evolve evaluation strategies to meet the demands of multi-step LLM systems can be demonstrated via a simple example of implementing Retrieval-Augmented Generation (RAG). RAG systems are conventionally judged by the final summary's fidelity to an ideal summary \cite{lin2004rouge, papineni2002bleu}. A more insightful evaluation, however, would be to determine the point of failure (e.g., retrieval vs. generation) and characterize its nature (e.g., irrelevant retrieval vs. incoherent retrieved sets), to provide a more accurate depiction of the system's capabilities and shortcomings. Suppose we extend this RAG system to include tools for retrieval and a reflection loop to self-correct those retrieved documents (e.g. \citet{yue2024inference}); every additional component that is added increases the scope of agent activity, the interaction effects, and the multitude of decisions that need to be tracked and evaluated. 

In this paper, we attempt to realize this evolution in evaluation capabilities. We compare different classes of offline evaluation architectures (Figure  \ref{fig:schematic}) with the explicit goal of detecting decisive, trajectory-altering faults in multi-step LLM systems. First, we structure the reasoning trajectories of the LLM-as-a-judge and introduce Evaluators to critique its rationales and final decisions. Second, we introduce iterative loops by which our Judge can reflect on the reasoning and confidence scores of the Evaluators and adjust its own assessments. We test these capabilities on the Who\&When benchmark \cite{zhang2025agent}, which consists of agentic logs and agent-step fault pairs derived from GAIA \cite{mialon2023gaia} and AssistantBench \cite{yoran2024assistantbench}. Our experiments demonstrate that structured iterative reasoning provides a significant and robust performance advantage over existing evaluation methods, and across diverse model families. Using Claude Sonnet 4, RAFFLES improved the step-level accuracy on fault attribution from a previous best of 38.10\% to 51.59\% on the Algorithmically-Generated dataset, and 18.20\% to 27.59\% on the Hand-Crafted dataset. We also demonstrated the versatility of RAFFLES on the ReasonEval benchmark \cite{xia2025evaluating}, which contains step-level annotations of reasoning chains for Math problems. With Claude Sonnet 4, RAFFLES improved step-level accuracy from 73.58\% on a strong one-shot baseline to 84.91\% on MR-MATH-Invalid, and from 75.46\% to 83.78\% on MR-GSM8K-Original. Results on ReasonEval help validate the effectiveness of RAFFLES in pinpointing the initial point of failure in complex reasoning chains, a critical requirement as we scale to more and more complex multi-component systems. 

\section{Related Work}

\textbf{Reasoning and automated evaluators}. The automated evaluation of LLMs (and agentic systems) has been enabled by (a) the extensive work done by researchers to introduce the LLM-as-a-judge \cite{fu2023gptscore, li2023alpacaeval, liu2023g, zheng2023judging}, and (b) recent advances in the reasoning capabilities of LLMs \cite{lightman2023let, wei2022chain, wang2022self, yao2023tree}.   Enhancements of the LLM-as-a-judge include new methods that incorporate planning within the prompt \cite{saha2025thinking}, self-check reasoning \cite{miao2023selfcheck},  debate \cite{chan2023chateval}, and juries as judges \cite{verga2024replacing}. Our contribution introduces a unique evaluation architecture that incorporates numerous reasoning advances; we introduce a Judge and a set of Evaluators that incorporate structured reasoning via problem decomposition \cite{zhou2022least, khot2022decomposed}, natural language confidence estimation \cite{lin2022teaching, xiong2023can, tian2023just}, as well as iterative refinement \cite{shinn2023reflexion, madaan2023self}.

\textbf{Fault attribution in agentic systems}. Numerous evaluation frameworks \cite{zhuge2024agent, zhang2025agent, deshpande2025trail, chang2024agentboard, liu2025mcpeval, pan2024autonomous} have shown preliminary success evaluating agentic benchmark datasets like GAIA, SWEbench, and AssistantBench \cite{mialon2023gaia, jimenez2023swe, yoran2024assistantbench}. Frameworks that detect trajectory-breaking faults include (a) a multi-agent debugging tool \cite{epperson2025interactive}, (b) a prompt-customized LLM-as-a-judge implementation \cite{deshpande2025trail}, and (c) methods that horse race against the LLM-as-a-judge, such as multi-pass binary search and Step by Step implementations \cite{zhang2025agent}. Following the release of the Who\&When dataset \cite{zhang2025agent}, recent work falls into two main categories: (1) simulation-based methods that simulate partial or entire trajectories to determine faults \cite{ge2025introducing}, and (2) simulation-free methods \cite{yu2025correct, ma2025automatic}, which our paper mainly focuses on. 

\textbf{Fault attribution in reasoning chains}. With the growing popularity of chain-of-thought \cite{wei2022chain, kojima2022large} and reasoning models (e.g. \citet{agarwal2025gpt}), advancement in pinpointing errors within a model's reasoning chain has been propelled by two concurrent efforts: the development of benchmarks with fine-grained, step-level error annotations \cite{xia2025evaluating, song2025prmbench, he2025can, zheng2024processbench}, and the proposal of novel detection methods. Novel methods include both training-free approaches \cite{hao2024llm} and fine-tuned Process Reward Models \cite{xia2025evaluating, she2025rprmreasoningdrivenprocessreward, zhang2025lessons}. We benchmark our framework on fault attribution within reasoning chains as a means to validate its performance and versatility.

\section{Background}
\label{sec:background}

A full trajectory from an agentic system or a step-by-step reasoning trajectory $\tau$ can be written as $\tau = (\tau_0, \ldots, \tau_T)$ where $T$ is a terminal time step. The success or failure of a trajectory is determined by a binary outcome function $Z(\tau)$: 
\begin{equation*}
\label{eq2_revised}
Z(\tau)=
\begin{cases}
1 & \text{if }\tau\text{ leads to the correct outcome},\\
0 & \text{otherwise}.
\end{cases}
\end{equation*}
In practice, $Z$ usually represents an external evaluator, such as an LLM-based judge, that evaluates the output of the pipeline $\tau$, assessing whether the final outcome meets the desired objective \cite{xie2024travelplanner, mialon2023gaia}. Given a dataset, we assume that there exists one or more $\tau^*$ such that $Z(\tau^*)=1$. When a system ultimately fails ($Z(\tau)=0$), we seek to understand where and when the failure occurs within the trajectory. 
Even when the final outcome is correct, there is still interest in understanding if and when there are errors in the reasoning process \cite{xia2025evaluating}.

To this end, we establish a precise hierarchy of faults to progress from local, procedural errors to the single, significant event that led to the outcome's failure.

\begin{definition}[\textbf{Step-Level Fault}] A step-level fault occurs at time $t$ if the step $\tau_t$ is deemed incorrect by a local judge function $f$ that takes in $(\tau_t)$. Specifically, $f$ produces a probability $c\in[0,1]$ representing the likelihood of error. A fault exists if $c = f(\tau_t) > \varepsilon$, where $\varepsilon$ is a predetermined threshold.
\label{def:step-fault}
\end{definition}

The function $f$ is usually a case-by-case LLM-judge that follows a certain taxonomy, most commonly hallucination or execution error \cite{deshpande2025trail, cemri2025multi}. 

\begin{definition}[\textbf{First Step-Level Fault Attribution}]
\label{def:first-step-fault}
We define the task of First Step-Level Fault Attribution as finding the first error that occurs in a multi-step LLM system. Specifically,
$$
\begin{aligned}
    t^* = \;& \underset{t}{\operatorname{arg\,min}} \quad t \\
    \text{subject to} \quad & (a) \:  f(\tau_t) > \varepsilon 
\end{aligned}
$$
\end{definition}
Definition \ref{def:first-step-fault} is helpful in cases where researchers want to identify the first failure step of the trajectory, such as when analyzing mathematical reasoning trajectories in ReasonEval \cite{xia2025evaluating}.

However, in cases where the trajectory ends up in a failure ($Z(\tau)=0$), a step-level error may or may not be directly responsible for the final failure. To establish this relationship, we must determine if a step-level fault had a tangible impact on the trajectory's result. To this end, we define two \textit{mutually exclusive} fault concepts.

\begin{definition}[\textbf{Trivial Fault}]
Let $\tau$ be a trajectory with a failed final outcome ($Z(\tau)=0$) containing a step-level fault $a_t$ such that no alternatives replacing $a_t$ with $\tilde a_t$ can result in a successful trajectory. Formally, $a_t$ is a trivial fault if 
$$
    Z(\tau)=0 \quad \land \quad \forall \tilde a_t \text{ s.t. } Z(\tau | a_t:=\tilde a_t) = 0
$$
\end{definition}
Here $\tau | a_t:=\tilde a_t$ denotes the alternative trajectory that unfolds after the intervention at time t. A fault is trivial when it was minor, corrected later, or if a subsequent, unavoidable fault later in the trajectory would have caused a failure regardless. Unlike a trivial fault, which has no bearing on the final outcome of a trajectory, we now define a critical fault that directly impacts the result.

\begin{definition}[\textbf{Critical Fault}]
\label{def:critical-fault}
Let $\tau$ be a trajectory with a failed final outcome ($Z(\tau)=0$) containing a step-level fault $a_t$ such that an alternative that replaces $a_t$ with $\tilde a_t$ results in a successful trajectory. Formally, $a_t$ is a critical fault if 
$$
    Z(\tau)=0 \quad \land \quad \exists \tilde a_t \text{ s.t. } Z(\tau | a_t:=\tilde a_t) = 1
$$
\end{definition}
This definition isolates actions that were decisively responsible for the failure.


Based on these concepts, we finally define a \textbf{Decisive Fault} as the \textit{first critical fault} in the trajectory. It is the earliest point at which an intervention would have salvaged the outcome, making it the primary origin of the failure. In the annotation guide in Who\&When benchmark \cite{zhang2025agent}, annotators were asked to first think of the most serious mistaken agent, implicitly considering the fault's severity. However, prior definitions of fault attribution have not formally accounted for this concept. To address this gap, we extend the definition of fault attribution by incorporating a step-level function, $f$, to quantify the severity of a fault.

\begin{definition}[\textbf{Decisive Fault Attribution}]
\label{def:decisive-fault}
The decisive fault is the first critical fault to occur in a trajectory. The process of identifying it can be formulated as a constrained optimization problem:
$$
\begin{aligned}
    t^* = \;& \underset{t}{\operatorname{arg\,min}} \quad t \\
    \text{subject to} \quad & (a) \: f(\tau_t) > \varepsilon \\
    & (b) \: Z(\tau | a_t := \tilde{a}_t) = 1, \exists \tilde{a}_t \in \mathcal{A}
\end{aligned}
$$
\end{definition}
\label{eq:decisive_fault}
Definition \ref{def:decisive-fault} refines the theoretical construct of a decisive fault from Who\&When by stipulating that it must be a non-trivial, high-severity error. This revised definition not only aligns more closely with the annotation guidelines for Who\&When but also motivates the design of our RAFFLES framework.

\section{RAFFLES}
\label{sec:raffles-primary}
Given a trajectory $\tau$, we designed RAFFLES (Algorithm \ref{alg:alg}) as a Judge-Evaluator offline iterative LLM system to attribute faults based on Definition \ref{def:decisive-fault}, which we distill into the following three necessary criteria:
\begin{enumerate}
    \item \textbf{Fault Condition}. The action of agent $i$ at step $t$ must constitute a step-level fault, satisfying Definition \ref{def:step-fault}.
    \item \textbf{Primacy}. The action of agent $i$ at step $t$ must be the earliest critical fault, satisfying the goal of finding the $\min$ of $t$ in Definition \ref{def:first-step-fault}.
    \item \textbf{Decisiveness}. The agent $i$'s mistake must be a critical fault, satisfying constraint (b) in Definition \ref{def:critical-fault}.
\end{enumerate}
The design of the Judge and Evaluators is specifically tailored to facilitate structured reasoning of these three criteria with confidence.

\textbf{Judge:} Judge $J$ receives relevant execution logs \(\tau\) and proposes the most likely candidate for the decisive fault, consisting of the candidate step $t$ and 3 criteria-driven rationales \(R_j=\{r_j^{1},r_j^{2},r_j^{3}\}\) for its candidate selection.  To enable structured reasoning, Judge provides rationales separately for the three criteria of decisive faults. The Judge is also asked to reference the log as needed for clarity, so as to improve the faithfulness of the rationale. See details of the Judge prompt in Appendix \ref{sec:appendix-raffle-prompt}. 

\textbf{Evaluator:} Each Evaluator $E_p, \text{ for } p\in\{1,2,3\}$ evaluates one of the three criteria for decisive faults given the proposed step $t$, \(\tau\), and the rationale $r_j^p$ from the Judge. Each $E_p$ returns a rationale $r_e^p$ assessing the criterion-specific soundness of the Judge's rationale $r_j^p$ and a confidence score $c_e^p$ between 0 and 100 to produce the likelihood of error in Definition \ref{def:step-fault}. We also designed an additional rule-based Evaluator $p=4$ to validate whether the proposed candidate step $t$ is consistent with the log $\tau$. The sum of confidences is denoted as $C = \sum_p c_e^p$ and the rationales are denoted as $R_e=\{r_e^p;p=\{1,2,3,4\}\}$. See the Evaluator prompt in Appendix \ref{sec:appendix-raffle-prompt}. 


To facilitate the iterative reasoning, the output of the Evaluators is appended to a memory component $\mathcal{H}$. $\mathcal{H}$ is then fed back to the Judge in the subsequent iteration, enabling it to refine its candidate step $t$ selection strategy. 


The iteration concludes when either (a) \textsc{Terminate}: $C$ is greater than a threshold of 350 after considering the log $\tau$ and the evidence in \(\mathcal{H}\), or (b) when a predefined maximum number of iterations \(K\) is reached and we take the candidate with the highest confidence in $\mathcal{H}$. The step $t$ that has the highest confidence $C$ in history $\mathcal{H}$ after this iteration is then considered the decisive fault $t^*$. If the Judge determines that there is no decisive fault, and the confidence from the Evaluators reaches the threshold, then $t^* =\text{None}$ signals an absence of decisive fault. 
An illustrative example of the iterative process can be found in Appendix \ref{sec:appendix-example}. 

Overall, RAFFLES relies on the iterative and structured reasoning process between the Judge and the Evaluators, tailored to solve the constraint optimization problem defined in Definition \ref{def:decisive-fault}. Given the challenges imposed by long-horizon, multi-component systems, we equip RAFFLES with more capabilities to increase overall evaluation accuracy. We introduce a Judge that uses structured reasoning to predict decisive faults, while a set of Evaluators check those predictions against unique criteria and report their agreement using natural language confidence estimates. The Judge then incorporates this feedback to iteratively refine decisive fault predictions to enable more accurate and reliable fault detection. In addition, RAFFLES prompts are task-agnostic; adding a new dataset only requires minor revisions to the problem statement. Details of our prompts can be found in Appendix \ref{sec:appendix-raffle-prompt}. 



\begin{algorithm}
  \caption{RAFFLES: Iterative Reasoning-based Fault Attribution}
  \label{alg:alg}
  \begin{algorithmic}[1]
    \REQUIRE Trajectory \(\tau\), max iterations \(K\)
    \STATE \(\mathcal{H}\gets\emptyset\) \COMMENT{Initiate memory component for history}
    \STATE \(k\gets 0\)
    \WHILE{not \(\textsc{Terminate}(\mathcal{H})\) and \(k\leq K\)}
      \STATE \(t, R_j \gets\mathrm{Judge}(\tau,\mathcal{H})\) \COMMENT{Propose fault step and rationale from \(\tau\)}
      \STATE \( C, R_e \gets\mathrm{Evaluator} (t,R_j,\tau)\) \COMMENT{Invoke Evaluator \(E_p\)}
      \STATE \(\mathcal{H} \gets \mathcal{H} \cup \{(C, R_e)\}\) \COMMENT{Update history}
      \STATE \(k\gets k+1\)
    \ENDWHILE
    \RETURN \(t^*\), the decisive fault identified from \(\mathcal{H}\)
  \end{algorithmic}
\end{algorithm}
\vspace{-3pt}

\section{Experiments}
\subsection{Baseline Methods}
\label{sec:baseline-methods}
We perform comprehensive experiments on RAFFLES with open-source and proprietary models of various model sizes and families, including Llama-3.3-70B-Instruct \cite{meta-llama-3-3-70b-instruct}, 
gpt-oss-20b model \cite{openai2025gptoss120bgptoss20bmodel}, and claude-sonnet-4-20250514 \cite{anthropic_claude_sonnet_4}, with results for additional models in Appendix \ref{sec:appendix-extended-results}. All of the open-source models are deployed on AWS p4de nodes with NVIDIA A100 80GB GPUs using vLLM \cite{kwon2023efficient} without quantization. For all experiments except Claude, a greedy search is used with a maximum context length of 128k tokens; Claude uses a maximum context length of 200K tokens. 

For a trajectory of an average 9 agentic turns, RAFFLES only cost roughly \$0.1188 using claude-sonnet-4-20250514 with an average latency of 15.55 seconds. The full cost and latency analysis of RAFFLES can be found in Appendix \ref{sec:appendix-latency}, where we detail the fractional cost of RAFFLES compared to today's reliable alternative (e.g., junior engineers) and how the latency of RAFFLES was reduced by over 50\% using simple parallelized LLM calls. 

In addition to RAFFLES, we implement four baselines categorized by the evaluation system hierarchy shown in Figure \ref{fig:schematic}, covering prior work as well as a strong Tool-Caller baseline that we introduce. The first baseline is a one-shot \textbf{LLM-as-a-judge} from \cite{zhang2025agent}, which directly evaluates the entire log and problem statement to find the faulty step. The second class of baseline evaluators include LLM routers proposed in Who\&When \cite{zhang2025agent}, including a \textbf{Step by Step} method, which examines the log sequentially and stopping only when an error is detected, and a \textbf{binary search} method that iteratively determines if an error lies in the upper or lower half of the log to narrow the search. Finally, we introduce a \textbf{Tool-Caller} baseline that uses a planner to decide which step to investigate, calling an LLM judge with the log index (e.g., \texttt{<tool>judge(id=1)</tool>}) up to three times. The prompts used for these baselines are available in Appendix \ref{sec:prompt}. 

\subsection{Dataset}
Our primary experiments are conducted on two subsets from the \textbf{Who\&When} dataset \cite{zhang2025agent}, a benchmark specifically created for fault attribution in multi-agent systems. The dataset consists of agentic logs and agent-step fault pairs derived from GAIA \cite{mialon2023gaia} and AssistantBench \cite{yoran2024assistantbench}. We utilize its two subsets, without ground truth, to evaluate performance across different agentic setups: (a) the Algorithmically-Generated subset provides breadth with 191 unique agents across 126 logs (avg. 9 steps), and (b) the Hand-Crafted subset offers depth with 5 agents across 58 logs and an avg. of 51 steps. Despite one subset having the name "Hand-Crafted", both subsets are generated using real and dynamic agentic systems that reflect real-world challenges. These agents include an Orchestrator, Web Agent, Coding Agent, and many more (see Appendix \ref{sec:appendix-dataset} Table \ref{tab:whowhen-frequent-faults}). 

To demonstrate RAFFLES' versatility, we tested fault attribution on mathematical reasoning using two datasets from \textbf{ReasonEval} \cite{xia2025evaluating}. The first dataset, MR-MATH-Invalid, contains 159 trajectories with correct output (avg. 9 steps). We use a subset of 83 trajectories that have an incorrect reasoning step in the experiments. The second dataset, MR-GSM8K, consists of 2,777 samples with question types that are "original" (avg. 7.05 steps) and "reversed" (avg. 11.26 steps). For ReasonEval tasks, the concept of a critical fault is not applicable based on how the dataset was annotated. Thus, to adapt RAFFLES for these datasets, we omit the Decisiveness criterion (Section \ref{sec:raffles-primary}) from our framework while the other components remain unchanged. A detailed discussion of our dataset selection criteria, statistics, and potential limitations is available in Appendix \ref{sec:appendix-dataset}.

\subsection{Evaluation Metrics}

We propose two metrics for fault attribution. (1) \textbf{Strict Step-Level Accuracy} (or simply Step-Level Accuracy): This is the primary metric, the proportion of test cases where the system correctly predicts the exact step number $t$ of the fault within the trajectory $\tau$. This serves as our most stringent measure of performance, requiring precise identification of the failure point. (2) \textbf{Tolerant Step-Level Accuracy}: This secondary metric measures accuracy within a tolerance window of $s$ steps. A prediction $\hat t$ is considered correct if it falls within $s$ steps of the ground-truth fault step (i.e. $|\hat t - t_{gt}|\leq s$). This metric is a good indicator of practical utility, as pinpointing a fault to a small window (e.g., $s \leq 2$) is often sufficient to guide efficient manual inspection and debugging of interacting components. We discuss alternative metrics for each dataset in Appendix \ref{sec:appendix-metrics}, and show experiment results using alternative metrics in Appendix \ref{sec:appendix-extended-results}. 

\section{Results}

\begin{table*}[!h]
\centering
\caption{Step-level accuracy on Who\&When datasets. Bold text highlights the best performance, and underlined text represents the second best. Alg. represents Algorithmically-Generated; Hand. represents the Hand-Crafted dataset. }
\label{tab:who-when-results}
\begin{tabular}{l cc cc cc}
\toprule
& \multicolumn{2}{c}{\textbf{Llama 3.3 70B}} & \multicolumn{2}{c}{\textbf{gpt-oss-20b}} & \multicolumn{2}{c}{\textbf{Claude Sonnet 4}} \\
\cmidrule(lr){2-3} \cmidrule(lr){4-5} \cmidrule(lr){6-7}
\textbf{Method} & \textbf{Alg.} & \textbf{Hand.} & \textbf{Alg.} & \textbf{Hand.} & \textbf{Alg.} & \textbf{Hand.} \\
\midrule
\textbf{Chat-LLM} & 19.05 & 6.90 & 14.29 & 3.45 & 29.37 & 3.45 \\
\textbf{Routers} (Step by Step) & 6.35 & 3.45 & 19.84 & \textbf{22.41} & 30.16 & 12.07 \\
\textbf{Routers} (Binary Search) & 4.76 & 10.34 & 15.08 & 15.52 & \underline{33.33} & \underline{20.69} \\
\textbf{Tool-Caller} & \underline{33.33} & \underline{13.56} & \underline{29.37} & \underline{17.24} & 30.95 & 18.97 \\
\textbf{RAFFLES} & \textbf{43.65} & \textbf{20.69} & \textbf{44.44} & \textbf{22.41} & \textbf{51.59} & \textbf{22.41} \\
\bottomrule
\end{tabular}
\end{table*}


\begin{table*}
 \centering
 \caption{Step-level accuracy on ReasonEval datasets. Baseline Chat-LLM is based on the prompt from \citet{xia2025evaluating}. Bold text highlights the best performance, and underlined text represents the second best. Math represents MR-MATH-Invalid dataset, gsm-o represents MR-GSM8K-Original, and gsm-r represents MR-GSM8K-Reversed.}
 \label{tab:reason-eval-results}
 \begin{tabular}{l ccc ccc ccc}
 \toprule
 & \multicolumn{3}{c}{\textbf{Llama 3.3 70B}} & \multicolumn{3}{c}{\textbf{gpt-oss-20b}} & \multicolumn{3}{c}{\textbf{Claude Sonnet 4}} \\
 \cmidrule(lr){2-4} \cmidrule(lr){5-7} \cmidrule(lr){8-10}
 \textbf{Method} & \textbf{math} & \textbf{gsm-o} & \textbf{gsm-r} & \textbf{math} & \textbf{gsm-o} & \textbf{gsm-r} & \textbf{math} & \textbf{gsm-o} & \textbf{gsm-r} \\
 \midrule
 \textbf{Chat-LLM} & 65.41 & 69.75 & 61.07 & \underline{84.28} & 84.77 & \textbf{78.44} & 73.58 & 75.46 & 67.77 \\
  \textbf{RAFFLES} $K=0$ & 68.55 & 74.68 & \textbf{62.18} & 83.02 & \underline{84.84} & 76.60 & \underline{82.39} & 82.65 & 70.42 \\
  \textbf{RAFFLES} $K=1$ & \textbf{69.18} & \underline{75.81} & 61.81 & \underline{84.28} & \textbf{85.12} & 77.26 & 79.87 & \underline{83.00} & \underline{74.83} \\
 \textbf{RAFFLES} $K=2$ & \underline{68.55} & \textbf{76.02} & \underline{61.95} & \textbf{84.91} & \textbf{85.12} & \underline{77.41} & \textbf{84.91} & \textbf{83.78} & \textbf{75.28} \\
 \bottomrule
 \end{tabular}
 \end{table*}

\subsection{RAFFLES Results on Who\&When}
Table \ref{tab:who-when-results} details the step-level accuracy on the Algorithmically-Generated and Hand-Crafted datasets, using a maximum of $K=2$ iterations. 

\textbf{RAFFLES detects decisive faults better than baseline methods across diverse model families.} On the Algorithmically-Generated dataset, RAFFLES outperforms our strongest baselines, Tool-Caller, by 31\% with Llama 3.3 70B and Binary Search, by 55\% with Claude Sonnet 4. The performance superiority is further validated on the more challenging Hand-Crafted dataset, where RAFFLES surpasses Tool-Caller by 53\% (Llama 3.3 70B). Furthermore, as shown in Table \ref{tab:accuracy-reordered}, RAFFLES consistently outperforms our Chat-LLM baseline across nearly all tolerance thresholds in both datasets. Together with the results of additional models in Appendix \ref{sec:appendix-extended-results}, we show that RAFFLES provides a universal performance lift across all benchmark models, from large-scale Mixture-of-Experts (MoE) models with pre-trained reasoning capabilities to smaller models with limited pre-trained reasoning. 

\begin{table*}
\centering
\caption{Tolerant step-level accuracy with $\pm s$ steps error tolerance using Llama 3.3 70B model.}
\label{tab:accuracy-reordered}
\begin{tabular}{ll cccccc}
\toprule
Dataset & Method & Acc. & \textbf{$\pm 1$} & \textbf{$\pm 2$} & \textbf{$\pm 3$} & \textbf{$\pm 4$} & \textbf{$\pm 5$} \\
\midrule
\textbf{Algorithmically-} & Chat-LLM (One Shot) & 19.05 & 49.21 & 64.49 & 75.40 & 86.51 & 90.48 \\
{\textbf{Generated}}& RAFFLES (Iterator) & \textbf{43.65} & \textbf{58.73} & \textbf{73.81} & \textbf{82.54} &\textbf{91.27} &\textbf{ 92.86} \\
\midrule
\multirow{2}{*}{\textbf{Hand-Crafted}} & Chat-LLM (One Shot) & 6.90 & 12.07 & 20.69 & \textbf{31.03} & 32.76 & 44.83 \\
& RAFFLES (Iterator) & \textbf{20.69} & \textbf{25.86} & \textbf{27.59} & 29.31 & \textbf{46.55} & \textbf{48.28} \\
\bottomrule
\end{tabular}
\end{table*}

\textbf{Fault attribution methods using partial trajectories are insufficient for robust performance improvement.} Performance with Step by Step appears to be highly dependent on the pretrained model (Table \ref{tab:who-when-results}) as powerful proprietary models overcome incomplete information, while other (open) models cannot. With a history limited to prior context, Step by Step fails on long trajectories like those in the Hand-Crafted dataset; e.g., accuracy drops from 6.9 to 3.45 for Llama 3.3 70B. The Binary Search method is moderately effective for long contexts, but less so for shorter ones; the greatest lift is for Claude-4-Sonnet on the Hand-Crafted dataset. Overall, these results highlight the need for a system that combines a global, end-to-end perspective with focused, local analysis. RAFFLES consistently outperforms these baselines by embodying these principles; the Judge operates on the global context to form hypotheses, and the Evaluators focus on the reasoning process for the hypothesized fault step.

\textbf{Structured reasoning is more effective than flexible tool calling in fault attribution. } While Table \ref{tab:who-when-results} shows that the Tool-Caller baseline introduces an improvement over the Router and Chat-LLM baselines, it is still consistently outperformed by RAFFLES. We attribute this performance gap to the lack of procedural reliability and robust reasoning within the Tool-Caller's Planner. Although designed for flexible and efficient decision-making, the Planner frequently fails to generate high-quality candidates for the Judge component to evaluate. 

\begin{figure}
\centering
\includegraphics[width=0.49\linewidth]{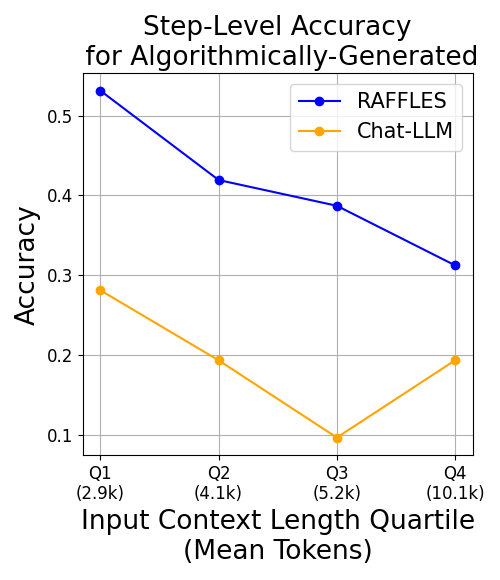}
\includegraphics[width=0.49\linewidth]{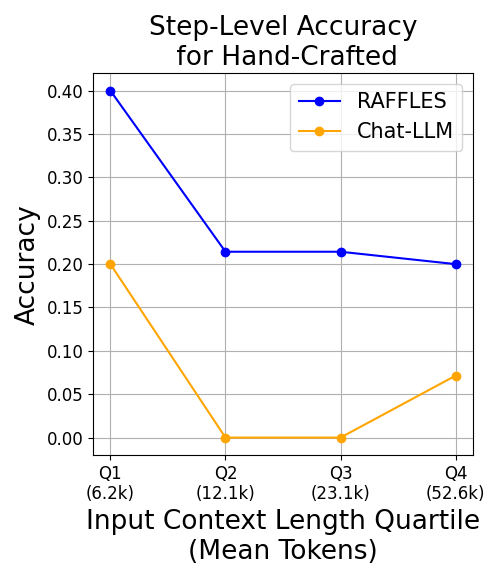}
\caption{
Llama 3.3 70B performance degrades with input trajectory length. Quartile of token length is with respect to the corresponding dataset.}
\label{fig:iteration_confidence_left}
\end{figure}

\textbf{RAFFLES consistently outperforms the baseline method on challenging long trajectories. }
While a global context is necessary for robust fault attribution, our analysis reveals a central dichotomy: the effectiveness of these methods is negatively impacted by increasing trajectory length. As Figure \ref{fig:iteration_confidence_left} reveals, a clear pattern emerges where step-level fault detection accuracy decreases as trajectories increase in context token length. This observation is directly attributable to the broader challenge models face in accurately pinpointing a decisive fault within long, complex trajectories. Despite this general degradation, RAFFLES still consistently outperforms single-LLM baselines across all context lengths. Moreover, Table \ref{tab:who-when-results} shows that on long-horizon agentic logs averaging 50 steps, RAFFLES improves performance from a 3.45 baseline to 22.41 for Claude and from a 6.90 baseline to 20.69 for Llama 3.3 70b, showcasing the strong performance of RAFFLES on long trajectories. 


\begin{table*}[ht!]
\centering
\caption{State-of-the-art step-level accuracy results on the Who\&When dataset without ground truth, as reported in recent literature. 
Bold text shows overall best performance, and underline highlights best performance per category.}
\begin{tabular}{lllccc}
\toprule
\textbf{Category} & \textbf{Methods} & \textbf{Model} & \textbf{Algo.} & \textbf{Hand.} \\
\midrule
Requires Simulation
& FAMAS \cite{ge2025introducing} & Qwen2.5-72B &  \underline{23.81} & \underline{\textbf{41.38}} \\
\midrule
\multirow{3}{*}{No Simulation} & CORRECT \cite{yu2025correct} & GPT-5 &  38.10 & 17.20 \\
& Causal Framework \cite{ma2025automatic} & N/A &  36.20 & 18.20 \\
& RAFFLES (Ours, $K=5$) & Claude Sonnet 4 & \textbf{\underline{51.59}} & \underline{27.59} \\
\bottomrule
\end{tabular}
\label{tab:who-when-sota}
\end{table*}

\textbf{Iterative verification limits error propagation in RAFFLES.}
A key concern of iterative agentic systems is the risk of early mistakes cascading across multiple iterations. RAFFLES mitigates this through the Judge–Evaluator design, where each attribution is repeatedly verified until all Evaluators reach high confidence score. As shown in Table \ref{tab:reason-eval-gsm8k-full} in the Appendix, accuracy consistently improves across iterations, and Table \ref{tab:iteration_comparison} in the Appendix shows substantial candidate turnover, reflecting active correction rather than passive carryover.
To assess the frequency of true error propagation, we measured cases where an incorrect prediction persists across all $K=2$ iterations. These cases occur in around 20.6\% of Algorithmically-Generated examples and 17.2\% of Hand-Crafted examples, indicating the persistent errors caused by consistently overconfident Evaluators. These patterns confirm that RAFFLES continuously revises its hypotheses rather than reinforcing early noise, thus limiting error propagation through iterations. 

\textbf{RAFFLES establishes a new state of the art on the Who\&When dataset} (Table \ref{tab:who-when-sota}). 
RAFFLES significantly outperforms the best simulation-free method by 35\% (Algorithmically-Generated) and 52\% (Hand-Crafted). More importantly, it also surpasses leading simulation-based methods by 8.7\% on the Algorithmically-Generated dataset. 
Our results demonstrate that structured iterative reasoning can supplant the need for costly simulations, particularly in shorter-horizon scenarios. In Appendix \ref{sec:appendix-extended-results}, we show results on agent-level metrics, where RAFFLES maintains superior performance across all datasets. 

\subsection{RAFFLES Results on ReasonEval}

\textbf{RAFFLES demonstrates strong generalization to fault attribution in reasoning chains.} 
A key advantage of RAFFLES is task-agnostic structured reasoning. With minimal prompt changes from agentic fault attribution, the results on the ReasonEval dataset, presented in Table \ref{tab:reason-eval-results}, show that RAFFLES (with $K=2$) significantly outperforms the LLM-judge baseline. When implemented with the Claude Sonnet 4 model, our method improves upon the baseline by 15.40\% on MR-MATH-Invalid, 11.08\% on MR-GSM8K-Reversed, and 11.03\% on MR-GSM8K-Original. Even with no iteration $K=0$, a single-pass with a structured LLM judge, Claude-based RAFFLES achieves a step-level accuracy of 82.39\% (+12\%) on MR-MATH-Invalid set and 82.65\% (+9.5\%) on MR-GSM8K-Original dataset, outperforming the original prompt with simple chain-of-thoughts. The consistent results across all model families demonstrate that structured reasoning is inherently more effective. 
Moreover, as detailed in Appendix \ref{sec:appendix-extended-results}, this performance surpasses previously reported results from models trained on the same dataset. These results validate the effectiveness of RAFFLES in pinpointing the initial point of failure in complex reasoning processes, a critical requirement as we scale to more complex multi-component systems. 

\textbf{Iterative reasoning improves fault attribution performance.} Table \ref{tab:reason-eval-results} shows RAFFLES performance when choosing different max step level $K$. We observed that iterative reasoning improves step-level accuracy for the three models at $K=2$ for almost all datasets. We found that (1) iterative reasoning helps LLMs narrow in on the portion of the trajectory most likely to contain faults, as evidenced by the convergence of the target and predicted distributions for fault steps in Figure \ref{fig:histogram-convergence}; (2) the efficacy of iterative reasoning is affected by the length of trajectory (Figure \ref{fig:right_column}), especially when the decisive fault is located in the middle of a trajectory, as observed in \citet{liu2023lost}; (3) iterative reasoning improvements are not always monotonic, echoing the findings of \citet{madaan2023self}, necessitating the design of high-confidence early stop criteria and a maximum step size $K$ (Table \ref{tab:reason-eval-results}). In effect, our experiments have shown that models benefit from structured iterative reasoning during fault attribution, and this has enabled our framework to outperform strong baselines. Further analysis on this topic can be found in Appendix \ref{appendix:iterative-reasoning}. 




\section{Conclusion}


We introduced RAFFLES, a novel architecture for post-hoc iterative evaluation, designed to systematically identify decisive faults in agentic systems. Our Judge-Evaluator iterative architecture builds a history of hypotheses and iteratively refines its reasoning, not only evaluating a candidate fault but also critiquing the reasoning behind its selection. By allowing error correction and iterative structured reasoning, which are central for problems that cannot be solved in a single step, our results show a step-level accuracy of over 51\% on the Algorithmically-Generated (compared to published simulation-free best of 38\%) and over 27\% on the Hand-Crafted dataset (versus best of 18\%). Finally, we demonstrated RAFFLES' versatility on fault attribution in math reasoning trajectories, showing 15.4\% improvement over baseline. 

\section*{Limitations}
\label{sec:appendix-limitation}

A primary, systemic limitation confronting the field is the pronounced scarcity of large-scale, high-quality datasets for agentic fault attribution. Our experiments were conducted on the publicly available Who\&When dataset, but the current data landscape is insufficient for the comprehensive validation and generalization of emerging methods. A critical direction for future research is the curation of larger and more diverse public benchmarks that can support more rigorous model evaluation. Concurrently, the development of reliable methodologies for generating high-fidelity synthetic data tailored for fault attribution tasks would represent a significant contribution to advancing the field.

While we demonstrated the applicability of RAFFLES to the ReasonEval dataset, our primary focus was not on creating a superior process reward model. This scope presents a clear avenue for future investigation: applying our methodology more broadly to fault attribution for reasoning models and systematically evaluating its performance against specialized reward-based approaches. Although the iterative reasoning employed by RAFFLES incurs higher latency than a fine-tuned process reward model, it remains highly viable for less time-sensitive applications. These include the in-depth analysis of reasoning processes for model evaluation and the generation of high-quality synthetic datasets for training more efficient downstream models.

\section*{Ethical Considerations}
This work has undergone a comprehensive internal risk evaluation. We have verified that all prior works, datasets, and models are properly cited and used in accordance with their licenses and intended purposes. All datasets were anonymized, and detailed statistics and descriptions can be found in the original source papers and in Appendix~\ref{sec:appendix-dataset}. The scope of this study is limited to English-language datasets with a focus on agentic systems and mathematical problems. We utilized an AI assistant for grammar checking, LaTeX formatting, and fluency improvements.

\bibliography{custom}


\newpage
\section*{Appendix}
\appendix

\section{Discussion on Datasets}
\label{sec:appendix-dataset}

As detailed in Table \ref{tab:dataset-stats}, a statistical comparison reveals that the Who\&When dataset presents a substantially more complex challenge than ReasonEval. The primary distinction lies in the trajectory length; all three ReasonEval subsets contain average trajectories of fewer than 200 words. In stark contrast, the Algorithmically-Generated and Hand-Crafted subsets of Who\&When are significantly longer, averaging 1,500 and 7,460 words, respectively. While the average number of steps in ReasonEval is comparable to the Who\&When Algorithmically-Generated subset, the exceptional length and intricacy of the Hand-Crafted subset establish it as the most demanding benchmark in our experiments.

\begin{table}
\centering
\small 
\setlength{\tabcolsep}{2pt} 
\caption{Basic statistics of the Who\&When and ReasonEval datasets.}
\label{tab:dataset-stats}
\begin{tabular}{lc ccc}
\toprule
\textbf{Dataset} & \textbf{Subset} & \textbf{\makecell{Sample\\Size}} & \textbf{\makecell{Avg.\\Steps}} & \textbf{\makecell{Avg.\\Words}} \\
\midrule
\multirow{2}{*}{\textbf{Who\&When}} & \makecell{Algorithmically-\\Generated} & 126 & 8.72 & 1,507.33 \\
\cmidrule{2-5}
& Hand-Crafted & 58 & 50.60 & 7,459.91 \\
\midrule
\multirow{3}{*}{\textbf{ReasonEval}} & MR-MATH-Invalid & 159 & 6.78 & 179.91 \\
\cmidrule{2-5}
& \makecell{MR-GSM8K-\\original} & 1,418 & 7.05 & 104.82 \\
\cmidrule{2-5}
& \makecell{MR-GSM8K-\\reversed} & 1,359 &  11.26 & 164.98\\
\bottomrule
\end{tabular}
\end{table}

\begin{table}[H]
\centering
\caption{Most frequent faults in the Who\&When dataset by generation method.}
\label{tab:whowhen-frequent-faults}
\setlength{\tabcolsep}{1pt} 
\begin{tabular}{c l r}
\toprule
\textbf{Source} & \textbf{Mistake Agent} & \textbf{Count} \\
\midrule
\multirow{5}{*}{\makecell{Algorithmically-\\Generated}} & Verification\_Expert & 18 \\
& PythonDebugging\_Expert & 7 \\
& DataAnalysis\_Expert & 6 \\
& Validation\_Expert & 5 \\
& DataVerification\_Expert & 5 \\
\cmidrule(l){1-3}
\multirow{4}{*}{\makecell{Hand-\\Crafted}} & WebSurfer & 33 \\
& Orchestrator & 18 \\
& Assistant & 4 \\
& FileSurfer & 3 \\
\bottomrule
\end{tabular}
\end{table}

\begin{table}[H]
\centering
\caption{Top-5 steps where the most mistakes occur in the Who\&When dataset}
\label{tab:mistake-steps-side-by-side}
\begin{tabular}{l r @{\hskip 1cm} l r}
\toprule
\multicolumn{2}{c}{\textbf{Algorithmically-Generated}} & \multicolumn{2}{c}{\textbf{Hand-Crafted}} \\
\cmidrule(r){1-2} \cmidrule(l){3-4}
\textbf{Step} & \textbf{Count} & \textbf{Step} & \textbf{Count} \\
\midrule
1 & 34 & 8 & 9 \\
0 & 20 & 4 & 9 \\
5 & 14 & 12 & 8 \\
3 & 13 & 9 & 3 \\
8 & 12 & 16 & 3 \\
\bottomrule
\end{tabular}
\end{table}

\subsection{Limitations of Who\&When Dataset}
The Who\&When dataset is one of the first open-source datasets to address the issue of fault attribution in agentic systems. It advances the field by giving a new perspective on evaluating language-conditioned agentic systems.  Each log in the dataset includes the initial problem statement, a complete trace of agent-to-agent communications, the ground truth answer of the initial problem, and an annotated label pinpointing the erroneous agent and the specific step number of the failure. We show the most frequent step-level and faults in Table \ref{tab:whowhen-frequent-faults} and Table \ref{tab:mistake-steps-side-by-side}, respectively. 

While the Hand-Crafted dataset has more steps on average (55 steps when compared to the Algorithmically-Generated average of 8 steps), there are only 5 possible agents across all steps and logs. Specifically, the WebSurfer agent is the decisive fault in 33 out of 58 instances. A trivial system that always guesses the WebSurfer can achieve at least 57\% accuracy. Another factor impacting agent-level detection in the Hand-Crafted dataset is that 97\% of the agent events are the WebSurfer and the Orchestrator. Epperson et al \cite{epperson2025interactive} found that there's some nuance to detecting decisive faults at an agent-level, given that on some occasions, they could only correct a faulty WebSurfer instance by fixing the \textit{preceding} planning step by the Orchestrator. This interactive nature between agents indicates that there is some subjectivity in assigning faults between planning and executing agents, and therefore, there may be some benefit to detecting agent- and step-level faults at some level of coarseness, or tolerance, rather than precisely at one step. 

In addition, we've also found that 3 cases out of 126 in the Algorithmically-Generated dataset and 3 cases out of 58 in the Hand-Crafted dataset have inconsistencies where the agent-step pair in the ground truth does not correspond to the actual agent-step pair in the log. Either the agent name in the ground truth is erroneous, or the step number is incorrect. The specific instances are \texttt{59.json}, \texttt{15.json}, \texttt{14.json} in Algorithmically-Generated dataset, and \texttt{22.json}, \texttt{20.json}, \texttt{49.json} in Hand-Crafted dataset. Fixing these erroneous cases is outside of the scope of this paper, but we advise future researchers to take notice that these erroneous cases exist.

\section{Alternative Evaluation Metrics}
\label{sec:appendix-metrics}
To ensure clarity and standardize our evaluation, we adopt step-level accuracy as the primary, unified metric across all experiments. While prior work has proposed other metrics for specific datasets, such as Who\&When \cite{zhang2025agent} and ReasonEval \cite{xia2025evaluating}, step-level accuracy provides the most consistent and stringent measure of performance for our purposes.

For instance, prior work on agentic trajectories has used an agent-level accuracy metric \cite{zhang2025agent}, which evaluates if the correct faulty agent is identified. However, our analysis of the Who\&When dataset reveals a significant agent-level label imbalance: the WebSurfer agent is the root cause of failure in 57\% of all trajectories. This skew allows a trivial baseline that always predicts this single agent to achieve 57\% accuracy, rendering the metric less informative. As a more fine-grained measure, step-level accuracy is therefore our primary focus. For completeness, we provide agent-level comparisons in Appendix \ref{sec:appendix-extended-results}.

Furthermore, the ReasonEval benchmark \cite{xia2025evaluating} employs step-level F1 and solution F1 scores. These metrics are necessary because the ReasonEval dataset contains both correct and incorrect trajectories, requiring models to first detect the presence of an error before localizing it and necessitating a separate solution-level metric. In contrast, Who\&When only contains failed trajectories, simplifying the task to error localization. To maintain a clear and unified evaluation framework, we use step-level accuracy in the main text. Nevertheless, we report results using the F1 metrics in Appendix \ref{sec:appendix-extended-results} to ensure direct comparison with the ReasonEval literature.

\section{Extended results on Who\&When Dataset and ReasonEval Dataset using alternative metrics}
\label{sec:appendix-extended-results}

\begin{table*}[!ht]
\centering
\caption{Full Results on the Who\&When dataset, comparing RAFFLES against baseline methods across various models.}
\label{tab:full-who-when-results}
\begin{tabular}{ll cc cc}
\toprule
& & \multicolumn{2}{c}{\textbf{\shortstack{Algorithmically- \\ generated}}} & \multicolumn{2}{c}{\textbf{\shortstack{Hand- \\ crafted}}} \\
\cmidrule(r){3-4} \cmidrule(l){5-6}
\textbf{Model} & \textbf{Method} & Step-level & Agent-level  & Step-level & Agent-level \\
\midrule

\multirow{5}{*}{\textbf{Llama 3.1 8B}} & \textbf{Chat-LLM}  & 15.08 & \textbf{53.97} & 6.90 & \textbf{62.07} \\
& \textbf{Routers} - Step by Step & 16.67 & 46.03 & 0.00 & 29.31 \\
& \textbf{Routers} - Binary Search & 7.14 & 23.81 & 5.17 & 50.00 \\
& \textbf{Tool-Caller}  & \underline{19.05} & 45.24 & \underline{18.97} & 53.45 \\
& \textbf{RAFFLES}& \textbf{34.92} & \underline{50.00} & \textbf{25.86}  & \underline{58.62}  \\

\midrule
\multirow{5}{*}{\textbf{Llama 3.3 70B}} & \textbf{Chat-LLM} & 19.05 & \underline{55.56} & 6.90 & \textbf{63.79}  \\
& \textbf{Routers} - Step by Step &  6.35&  23.02 & 3.45 & 17.24 \\
& \textbf{Routers} - Binary Search & 4.76 & 32.54 & 10.34 & 53.45 \\
& \textbf{Tool-Caller} & \underline{33.33} & 46.03 & \underline{13.56} & 50.85  \\
& \textbf{RAFFLES} & \textbf{43.65} & \textbf{61.90} & \textbf{20.69} & \underline{60.34}  \\

\midrule
\multirow{5}{*}{\textbf{Mixtral 8x22B}} & \textbf{Chat-LLM} & \underline{20.63} & \underline{35.71} & 1.75 & \underline{52.63} \\
& \textbf{Routers} - Step by Step & 10.32 & 28.57 & 0.00 & 3.45 \\
& \textbf{Routers} - Binary Search & 0.79 & 33.33 & \underline{10.34} & \textbf{53.45} \\
& \textbf{Tool-Caller} & 17.46 & 31.75 & {7.02} & 38.60 \\
& \textbf{RAFFLES}& \textbf{29.37} & \textbf{47.62} & \textbf{15.79} & 38.60 \\

\midrule
\multirow{5}{*}{\textbf{gpt-oss-20b}} & \textbf{Chat-LLM} & 14.29 & \textbf{63.49} & 3.45 & 25.86 \\
& \textbf{Routers} - Step by Step & 19.84 & 34.92& 22.41 & 58.62 \\
& \textbf{Routers} - Binary Search & 15.08 & 38.89 &  15.52 & \textbf{53.45} \\
& \textbf{Tool-Caller} & \underline{29.37} & 41.27 & \underline{17.24} & 39.66 \\
& \textbf{RAFFLES} & \textbf{44.44} & \underline{58.73} & \textbf{22.41} & \underline{53.45} \\

\midrule
\multirow{5}{*}{\textbf{Claude Sonnet 4}} & \textbf{Chat-LLM} & 29.37 & \underline{62.70} & 3.45 & \textbf{60.34} \\
& \textbf{Routers} - Step by Step & 30.16 & 38.89 & 12.07 & 27.59 \\
& \textbf{Routers} - Binary Search & \underline{33.33} & 52.38 & \underline{20.69} & 51.72 \\
& \textbf{Tool-Caller} & 30.95 & 41.27 & 18.97 & 51.72 \\
& \textbf{RAFFLES} &  \textbf{51.59} & \textbf{67.46} & \textbf{22.41} & \textbf{60.34} \\

\bottomrule
\end{tabular}
\end{table*}

\begin{table*}[!ht]
\centering
\caption{State-of-the-art agent-level and step-level results on the Who\&When dataset, as reported in recent literature. These results have not been independently reproduced by the authors. An entry of 'N/A' in the 'Model' column signifies that the method does not require the use of large language models. 'Agent' represents agent-level accuracy and 'Step' represents step-level accuracy. Bold text shows the best performance in each category.}
\label{tab:appendix-who-when-sota}
\setlength{\tabcolsep}{2pt} 
\begin{tabular}{lllcccc}
\toprule
\textbf{Category} & \textbf{Methods} & \textbf{Model} & \multicolumn{2}{c}{\textbf{\makecell{Algorithmically- \\ generated}}} & \multicolumn{2}{c}{\textbf{\makecell{Hand- \\ crafted}}} \\
\cmidrule(lr){4-5} \cmidrule(lr){6-7}
& & & \textbf{Agent} & \textbf{Step} & \textbf{Agent} & \textbf{Step} \\
\midrule
\multirow{3}{*}{Baseline} & All-at-once \cite{zhang2025agent}& GPT-4o & 51.12 & 13.53 & 53.44 & 3.51 \\
& Step-by-step \cite{zhang2025agent} & GPT-4o & 26.02 & 15.31 & 32.75 & 8.77 \\
& Binary Search \cite{zhang2025agent} & GPT-4o & 30.11 & 16.59 & 36.21 & 6.90 \\
\midrule
\multirow{2}{*}{\makecell{Requires\\Simulation}} & A2P \cite{west2025abduct}& gpt-oss-120b & 65.40 & 47.46 & 58.62 & 29.31 \\
& FAMAS \cite{ge2025introducing} & Qwen2.5-72B & 55.56 & 23.81 & 62.07 & \textbf{41.38} \\
\midrule
\multirow{3}{*}{No Simulation} & CORRECT \cite{yu2025correct} & GPT-5 & -- & 38.10 & -- & 17.20 \\
& Causal Framework \cite{ma2025automatic} & N/A & 48.50 & 36.20 & 56.80 & 18.20 \\
& \textbf{RAFFLES} (Ours) & Claude Sonnet 4 & \textbf{67.46} & \textbf{51.59} & \textbf{67.24} & 27.59 \\
\bottomrule
\end{tabular}
\end{table*}

In this section, we show results using alternative metrics and additional state-of-the-art methods to fully present the performance of RAFFLES. Notice that additional state-of-the-art methods have not been independently reproduced by the authors due to several factors, including computational resource limitations, lack of access to OpenAI closed-source models, or the absence of reproducible prompts in the original publications.

\subsection{Who\&When}
On the Who\&When dataset, Table \ref{tab:full-who-when-results} shows both agent-level and step-level results with a diverse choice of model families of Llama 3.3 70B, Llama 3.1 8B, Mixtral 8x22B, gpt-oss-20b, and Claude Sonnet 4. Across these different model families, RAFFLES consistently demonstrates superior performance. It keeps the top rank for step-level accuracy in almost every model on both the algorithmically-generated and hand-crafted data splits. For agent-level accuracy, it consistently ranks either first or a close second. These results highlight the robustness and effectiveness regardless of the underlying language model.

When compared with other state-of-the-art methods (Table \ref{tab:appendix-who-when-sota}), RAFFLES establishes new performance benchmarks. Our method achieves the highest agent-level accuracy on both the algorithmically-generated (67.46\%) and hand-crafted (67.24\%) test sets, outperforming all prior work, including those that require simulation. Furthermore, RAFFLES also sets a new state-of-the-art in step-level accuracy on the algorithmically-generated data with a score of 51.59\%. 

\subsection{ReasonEval}
Although not primarily designed as a process reward model, RAFFLES demonstrates notable adaptability to other tasks, specifically fault attribution in reasoning chains. As detailed in Table \ref{tab:math-reasoning-sota}, RAFFLES outperforms current state-of-the-art reported results on the ReasonEval dataset from recent literature. Our model achieves superior performance against both fine-tuned baselines, such as Llemma-34B, and models without task-specific fine-tuning, such as GPT-4. These results indicate that the RAFFLES framework can be effectively extended to fault attribution for reasoning chains.

\section{Latency and Cost Analysis of RAFFLES}
\label{sec:appendix-latency}
Table \ref{tab:who-when-latency} highlights a trade-off inherent to our methodology: the RAFFLES approach generally shows a higher average latency compared to the simple Chat-LLM baseline across all evaluated models and testing paradigms. This increased processing time is an expected consequence of our method's design, as its iterative, multi-step refinement process introduces additional inference calls compared to one-off LLM baselines. In a typical 2 full iteration RAFFLES run, there would be 2 Judge calls and 6 Evaluator calls sequentially. After parallelizing the Evaluator calls per iteration, the total latency is equivalent to around 4 LLM calls (reduced by half). Empirical results in Table \ref{tab:who-when-latency} show that the latency was reduced by over 50\% for the Claude model.  

With regards to the cost to run RAFFLES, based on statistics in Table \ref{tab:dataset-stats} and prompts provided in Appendix \ref{sec:appendix-raffle-prompt}, per iteration of RAFFLES (4 LLM calls) would require around 9.8k input tokens for the shorter Algo. subset (avg 9 agentic turns) and 41k input tokens for the longer Hand. subset(avg 50 agentic turns), with 2k output tokens for each. Given Claude’s pricing and 2 max iterations, this is equivalent to \$0.1188 per log of Algo., and \$0.306 per log of Hand. Compared to junior engineers whose hourly rate is approximately \$50 per hour in the US, and assuming that they take 3 minutes to debug 9 agentic turns and 10 minutes to debug 50 turns; this is equivalent to \$2.5 per log of Algo, and \$12.5 per log of Hand. Despite multiple LLM calls, RAFFLES has a very low cost with relatively high accuracy.

\begin{table*}[]
\centering
\caption{State-of-the-art Solution F1 and Step F1 results on the ReasonEval dataset, as reported in recent literature. These results have not been independently reproduced by the authors. Bold values indicate the best performance in each column. }
\label{tab:math-reasoning-sota}
\setlength{\tabcolsep}{2pt} 
\begin{tabular}{lcccccc}
\toprule
& \multicolumn{2}{c}{\textbf{MR-MATH-Invalid}} & \multicolumn{2}{c}{\textbf{MR-GSM8K-Original}} & \multicolumn{2}{c}{\textbf{MR-GSM8K-Reversed}} \\
\cmidrule(lr){2-3} \cmidrule(lr){4-5} \cmidrule(lr){6-7}
\textbf{Method} & \textbf{Solution F1} & \textbf{Step F1} & \textbf{Solution F1} & \textbf{Step F1} & \textbf{Solution F1} & \textbf{Step F1} \\
\midrule
\multicolumn{7}{l}{\textbf{\textit{Previous SOTA results, as reported in \citet{xia2025evaluating}}}} \\
GPT-4       & 73.2 & 61.0 & 81.7 & 69.0 & 72.2 & 52.2 \\
Math-shepherd-Mistral-7b    & 70.1 & 60.0 & 86.0 & 73.4 & 77.2 & 59.6 \\
ReasonEval WizardMath-7B-V1.1 & 78.6 & 73.9 & 74.1 & 72.8 & 74.4 & 70.5 \\ 
ReasonEval Llemma-34B    & 79.6 & 77.5 & 81.0 & 73.5 & 76.1 & 69.3 \\
\midrule
\multicolumn{7}{l}{\textit{\textbf{Our experiments}}} \\
RAFFLES Llama 3.3 70B   & 84.3 & 75.7 & 89.8 & 76.9 & 84.2 & 65.2 \\
RAFFLES Mixtral 8$\times$22B  & 73.3 & 64.2 & 72.2 & 59.7 & 71.0 & 54.1 \\
RAFFLES Claude Sonnet 4 & \textbf{92.4} & \textbf{86.7} & 95.1 & 82.3 & 91.7 & 70.7 \\
RAFFLES gpt-oss-20b     & 89.3 & 79.8 & \textbf{ 95.5}   & \textbf{84.2}   & \textbf{93.7}   & \textbf{73.1}   \\
\quad Compared to previous best	    & $+13.6\%$ &	$+3.0\%$ & 	$+11.0\%$	  & $+14.6\%$	& $+21.4\%$	& $+3.7\%$ \\
\bottomrule
\end{tabular}
\end{table*}

\begin{table*}
\centering
\begin{threeparttable}
\caption{Average Latency (s) on Who\&When datasets}
\label{tab:who-when-latency}
\begin{tabular}{l cc cc cc}
\toprule
& \multicolumn{2}{c}{\textbf{Llama 3.3 70B}} & \multicolumn{2}{c}{\textbf{gpt-oss-20b}} & \multicolumn{2}{c}{\textbf{Claude Sonnet 4}} \\
\cmidrule(lr){2-3} \cmidrule(lr){4-5} \cmidrule(lr){6-7}
\textbf{Method} & \textbf{Alg.} & \textbf{Hand.} & \textbf{Alg.} & \textbf{Hand.} & \textbf{Alg.} & \textbf{Hand.} \\
\midrule
\textbf{Chat-LLM} & 8.69 & 7.81 & 10.22 & 34.80 & 5.94 & 8.49\\
\textbf{Routers} (Step by Step) & 18.70 & 129.3 & 3.56 & 12.72 & 24.5 & 200.34 \\
\textbf{Routers} (Binary Search) & 20.16 & 32.98 & 3.54& 7.89 & 10.39 & 21.15 \\
\textbf{Tool-Caller} & 32.36 & 53.40 & 12.69 & 24.73 & 24.86 & 40.79 \\
\textbf{RAFFLES} & 64.33 & 48.42 & 37.47 & 49.88 & 38.97\tnote{a} & 64.18\tnote{a} \\
\bottomrule
\end{tabular}

\begin{tablenotes}
    \item[a] The latencies for Claude were further reduced from 38.97 to 15.55 (Alg.) and 64.18 to 29.15 (Hand.) by parallelizing LLM calls for Evaluators. Similarly, we expect around a 50\% reduction of latency to other benchmarked models for RAFFLES.  
\end{tablenotes}
\end{threeparttable}
\end{table*}

\begin{table}[H]
\centering
\caption{Percentage of changed candidate step $t$ between current and prior iterations}
\label{tab:iteration_comparison}
\begin{tabular}{l cc}
\toprule
\textbf{Iteration} & \textbf{\makecell{Algorithmically-\\Generated}} & \textbf{\makecell{Hand-\\Crafted}} \\
\midrule
1$\rightarrow$2 & 25.40\% & 37.93\% \\
2$\rightarrow$3 & 14.29\% & 8.62\% \\
3$\rightarrow$4 & 3.97\% & 3.45\% \\
4$\rightarrow$5 & 5.56\% & 5.17\% \\
\bottomrule
\end{tabular}
\end{table}

\section{Extended Analysis of Iterative Reasoning}
\label{appendix:iterative-reasoning}

\begin{figure*}
    \centering
    \begin{subfigure}[b]{\linewidth}
        \includegraphics[width=0.32\linewidth]{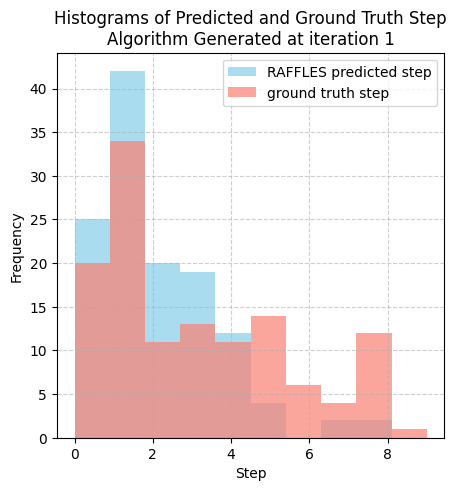}
        \includegraphics[width=0.32\linewidth]{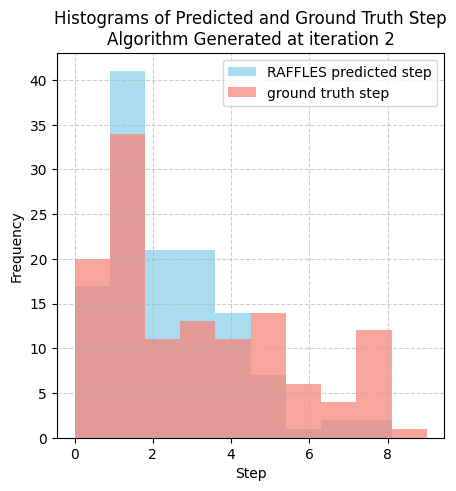}
        \includegraphics[width=0.32\linewidth]{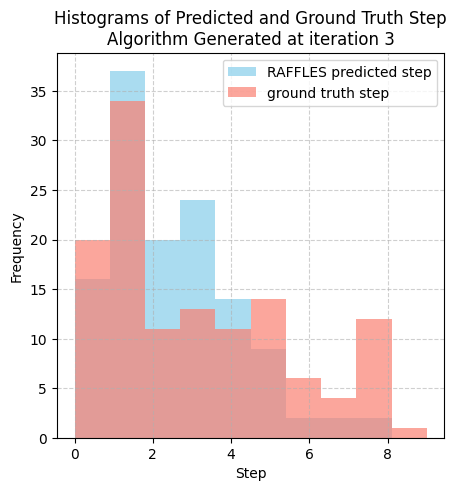}
        \caption{Histogram of predicted and ground truth step for Algorithmically-Generated dataset, from iteration 1 to 3. }
        \label{fig:algo-iteration}
    \end{subfigure}
    \begin{subfigure}[b]{\linewidth}
        \includegraphics[width=0.32\linewidth]{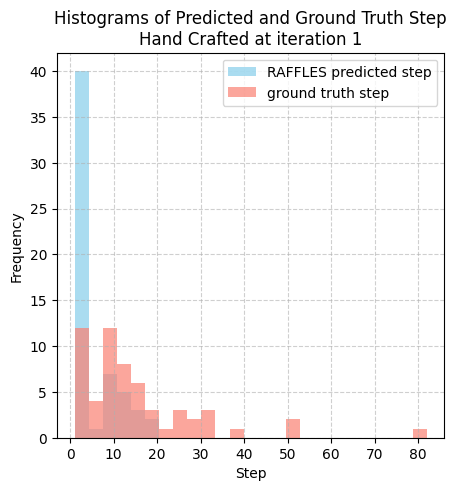}
        \includegraphics[width=0.32\linewidth]{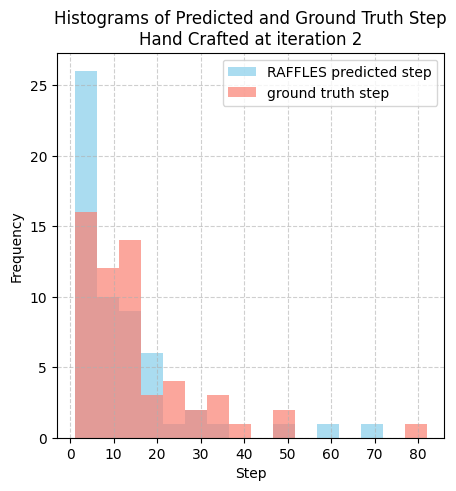}
        \includegraphics[width=0.32\linewidth]{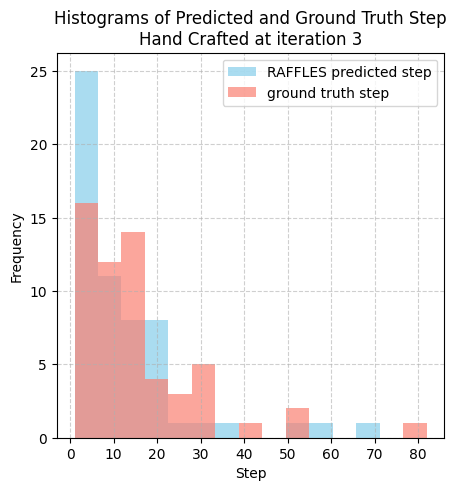}
        \caption{Histogram of predicted and ground truth step for Hand-Crafted dataset, from iteration 1 to 3. }
        \label{fig:hand-iteration}
    \end{subfigure}
    \caption{Histogram of predicted and ground truth step for Who\&When dataset, from iteration 1 to 3 for Llama 3.3 70B model. Notice that, for a given iteration $k$, we do not show the data points where convergence already happened in prior steps $k'<k$, hence the difference in ground truth histogram. }
    \label{fig:histogram-convergence}
\end{figure*}

Figure \ref{fig:histogram-convergence} illustrates the convergence from iteration 1 to 3 on the Who\&When dataset using Llama 3.3 70B. At an aggregate level, the predicted steps for both datasets converge toward the ground truth distribution. Focusing on the Hand-Crafted dataset in Figure \ref{fig:hand-iteration}, the model initially tends to predict the decisive fault at an early step ($\hat t \leq 5$), resulting in a prominent spike in the first iteration's distribution. Over subsequent iterations, this distribution converges towards that of the ground truth, showcasing the efficacy of the iterative process.

Similarly, in the ReasonEval GSM8K subset in Table \ref{tab:reason-eval-gsm8k-full}, the benefit of iterative reasoning is more apparent. There is a clear improvement in accuracy and F1 score as iteration increases from $K=0$ (only structured reasoning) to $K=2$ (maximum 2 iterations) on all of the subsets. This presents strong evidence that iterative reasoning works.

\begin{table*}
\centering
\caption{Accuracy and F1 results for ReasonEval GSM8K subsets using Claude Sonnect 4 model }
\label{tab:reason-eval-gsm8k-full}
\begin{tabular}{llcccc}
\toprule
                          &       & \textbf{Solution Acc.} & \textbf{Step Acc.} & \textbf{Solution F1} & \textbf{Step F1} \\
\midrule

\multirow{4}{*}{MR-GSM8K-Original} & judge & 87.80                  & 75.46                  & 87.80                  & 72.38                  \\
                          & K=0   & 94.36                  & 82.65                  & 94.34                  & 79.34                  \\
                          & K=1   & \textbf{95.13}         & 83.00                  & \textbf{95.13}         & 81.58                  \\
                          & K=2   & 95.06                  & \textbf{83.78}         & 95.06                  & \textbf{82.29}         \\ \midrule

\multirow{4}{*}{MR-GSM8K-Reversed} & judge & 85.65                  & 67.77                  & 85.35                  & 61.73                  \\
                          & K=0   & 87.86                  & 70.42                  & 87.46                  & 64.39                  \\
                          & K=1   & 91.61                  & 74.83                  & 91.49                  & 70.20                  \\
                          & K=2   & \textbf{91.83}         & \textbf{75.28}         & \textbf{91.71}         & \textbf{70.70}         \\ \midrule

\multirow{4}{*}{MR-GSM8K-POT}      & judge & 81.08                  & 64.86                  & 80.63                  & 67.12                  \\
                          & K=0   & 86.94                  & 72.97                  & 86.82                  & 74.67                  \\
                          & K=1   & 94.14                  & 80.18                  & 94.14                  & 83.84                  \\
                          & K=2   & \textbf{95.95}         & \textbf{83.33}         & \textbf{95.94}         & \textbf{86.20}         \\ \midrule

\multirow{4}{*}{MR-GSM8K-All}      & judge & 86.33                  & 71.19                  & 86.25                  & 66.07                  \\
                          & K=0   & 90.86                  & 76.39                  & 90.74                  & 70.68                  \\
                          & K=1   & 93.46                  & {79.09}      & 93.44                  & {75.56}      \\
                          & K=2   & \textbf{{93.66}} & \textbf{{79.89}} & \textbf{{93.64}} & \textbf{{76.28}} \\ \bottomrule

\end{tabular}
\end{table*}

\begin{figure}
    \centering
    \includegraphics[width=0.48\linewidth]{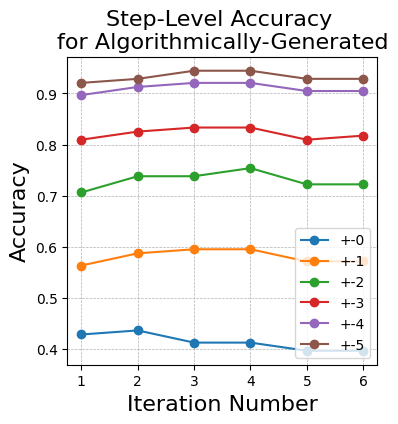}
    \includegraphics[width=0.48\linewidth]{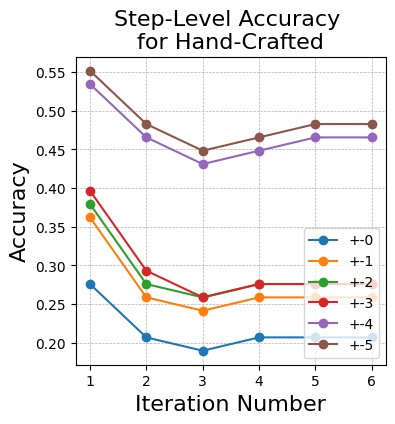}
    \caption{Llama 3.3 70B step-level accuracy vs. number of reasoning iterations}
    \label{fig:right_column}
\end{figure}

On the Who\&When dataset, an intriguing divergence of performance emerges in our experiments, as illustrated for the Llama 3.3 70B model in Figure \ref{fig:right_column}. On the Algorithmically-Generated dataset, the accuracy steadily improves from 70.35\% to 75.40\% with $\pm2$ tolerance by the fourth iteration. However, on the more challenging Hand-Crafted dataset, accuracy initially decreases before recovering and stabilizing at an equilibrium. This is likely caused by the Hand-Crafted dataset's larger and more complex search space, which impedes convergence. 

The primary distinction between the Algorithmically-Generated (Figure \ref{fig:algo-iteration}) and Hand-Crafted (Figure \ref{fig:hand-iteration}) datasets lies in their ground truth distributions. Most ground truth steps in the Algorithmically-Generated dataset are within the first four steps, while the Hand-Crafted dataset has a more distributed range of steps. The LLM's initial predictions often favor these early steps, a bias evident in Figure \ref{fig:hand-iteration}. We hypothesize this is analogous to the "lost in the middle" phenomenon reported for LLMs in long-context scenarios \cite{liu2023lost}. Furthermore, the "first mistake" framing in the prompt may predispose the model to focus on the initial part of the trajectory, further necessitating an iterative process with the "correction prompt" within the 3rd Evaluator's instructions. Beginning in iterations 2 and 3, critiques from this Evaluator prompt the Judge to reason more deeply and select candidates in later steps, thereby refining the prediction distribution to more closely match the ground truth.

This initial prediction bias, combined with the inherently larger search space of the Hand-Crafted problems, increases the reasoning difficulty. This confluence of factors contributes to the accuracy degradation observed in Figure \ref{fig:right_column}. Furthermore, we analyze the rate of convergence by computing the percentage of changed steps between consecutive iterations, presented in Table \ref{tab:iteration_comparison}. While more than a quarter of the steps are revised between the first and second iterations, the rate of change decreases substantially by iteration 4. This pattern indicates diminishing returns, a characteristic also observed in prior research on self-refinement methods \cite{madaan2023self}, and highlights the necessity of an early stopping mechanism, such as a maximum iteration count $K$, to cap the reasoning process.

\onecolumn
\section{Prompt}
\label{sec:prompt}
\subsection{Baseline prompt}
We took the one-shot prompt in \cite{zhang2025agent} and slightly modified the prompt to the Llama model chat template format. 
\begin{promptbox}{Chat-LLM Prompt}
\begin{verbatim}
You are an AI assistant tasked with analyzing a multi-agent 
conversation history when solving a real world problem.
The problem is: {problem}.
Identify which agent made an error, at which step, and explain the 
reason for the error.
Here’s the conversation: {failure_log}
Based on this conversation, please predict the following:
1. The name of the agent who made a mistake that should be directly 
responsible for the wrong solution to the real
world problem. If there are no agents that make obvious mistakes, 
decide one single agent in your mind. Directly
output the name of the Expert.
2. In which step the mistake agent first made mistake. For example, 
in a conversation structured as follows:
{{
”agent a”: ”xx”,
”agent b”: ”xxxx”,
”agent c”: ”xxxxx”,
”agent a”: ”xxxxxxx”
}},
each entry represents a ’step’ where an agent provides input. 
The ’x’ symbolizes the speech of each agent. If the
mistake is in agent c’s speech, the step number is 2. If the second 
speech by ’agent a’ contains the mistake, the step
number is 3, and so on. Please determine the step number where the 
first mistake occurred.
3. The reason for your prediction. Please answer in the format. 
Notice that you can *ONLY* output the json. 
```json
{{
    "agent_name": "",
    "step_number": 0,
    "reason_for_mistake": ""
}}
```
\end{verbatim}
\end{promptbox}

For both the Step by Step baseline and the binary search baseline, we use the following prompt template. We then use the specific task description to describe the required logic for Step by Step. These prompts are taken mostly from the Who\&When paper with slight modification \cite{zhang2025agent}. 

\begin{promptbox}{Prompt Template}
    \begin{verbatim}
You are an intelligent assistant that takes in a task 
description, and task output and complete based on 
requirements. 

Task Description
{task_description}

Input Metadata
{input_metadata}

Task Output
{task_output}

Remember, that your output should only be a json and nothing else. 
\end{verbatim}
\end{promptbox}

\begin{promptbox}{Step by Step instruction}
    \begin{verbatim}
You are an AI assistant tasked with evaluating the correctness of 
each step in an ongoing multi-agent conversation aimed at solving 
a real-world problem. Based on the conversation history in 
"history_up_to_step" up to the current step. 
Your task is to **determine whether the most recent agent’s action 
contains an error that could hinder the problemsolving process**. 
Please respond with ’Yes’ or ’No’ and provide a clear explanation for 
your judgment.
Note: Please avoid being overly critical in your evaluation.
Attention: Respond in the format:
1. Yes/No. Yes being that the pipeline failure is because of this 
latest step and No being that the  pipeline failure is not because of 
this step.
2. Reason for the judgment.
Remember that you are not trying to answer the question  based on the 
question given. Your job is to **determine  whether the most recent 
agent’s action contains an error  that could hinder the problemsolving 
process**. 
\end{verbatim}
\end{promptbox}

\begin{promptbox}{Step by Step task output}
    \begin{verbatim}
Please answer in the format:
```json
{
    "judgement": "upper half" or "lower half",
    "reason": ""
}
```
\end{verbatim}
\end{promptbox}

\begin{promptbox}{Binary Search instruction}
    \begin{verbatim}
You are an AI assistant tasked with analyzing a segment of a multi-agent 
conversation. Multiple agents are collaborating to address a user query, with 
the goal of resolving the query through their collective dialogue.
Your primary task is to identify the location of the most critical mistake within 
the provided segment. Determine which half of the segment contains the single step 
where this crucial error occurs, ultimately leading to the failure in resolving 
the user’s query.
The problem to address is as follows: {problem}
Review the following conversation segment {range_description}:

{chat_segment_content}

Based on your analysis, predict whether the most critical error is more likely to 
be located in the upper half ({upper_half_desc}) or the lower half 
({lower_half_desc}) of this segment.
Please simply output either 'upper half' or 'lower half'. You should not output 
anything else."""

    binary_search_task_output = """
Please answer in the format:
```json
{
    "judgement": "upper half" or "lower half",
    "reason": ""
}
```
\end{verbatim}
\end{promptbox}

\begin{promptbox}{Binary Search task output}
    \begin{verbatim}
Please answer in the format:
```json
{
    "judgement": "upper half" or "lower half",
    "reason": ""
}
```
\end{verbatim}
\end{promptbox}

\subsection{Tool-Caller prompt}

Our Tool-Caller mechanism is implemented using LangChain \footnote{https://www.langchain.com/} and is designed to improve upon traditional Step-by-Step methods. It features a \textbf{Planner} that leverages global access to the entire log to intelligently select a specific step to inspect via its id. A \textbf{Judge}, similar to the one in the Step by Step approach, then evaluates the process up to the step designated by the Planner.

This architecture allows the Tool-Caller to dynamically select the most relevant point for evaluation, breaking free from a predefined, rigid iterative sequence. Because the Planner has a global view of the process, its selection of the next candidate to inspect is significantly more efficient and intelligent.

Despite having a planner and judge like RAFFLES, the Tool-Caller method does not necessarily have a structured and iterative reasoning process. The planner does not reason explicitly before giving out the next candidate to check. There is only one judge, which is insufficient to provide enough reasoning to support the complexity of the fault attribution task. 

\begin{promptbox}{Tool-Caller Prompt}
\begin{verbatim}
You are an expert in planning 
and calling agents to evaluate the input. 
You are given a system log and a set of possible agents.
Based on the log, you will need to make one or more agent calls to 
achieve the purpose.
If none of the agents can be used, point it out. If the given question 
lacks the parameters required by the function, also point it out. 
If you decide to invoke any of the function(s), you MUST put it in the 
format of <agent>agent_name(args, kwargs)</agent>
You SHOULD NOT include any other text in the response. You should only 
call each agent ONCE. 
Here is a list of agents in JSON format that you can invoke.
1. You don't have to look over the entire conversation history one 
after another, it is okay to choose the most important one first. 
2. There is only one mistake in each conversation history, so you only 
need to find one agent and one step.
3. Your goal is to find the agent at fault with the least number of agent 
calls. So choose the agent to inspect wisely based on which agent looks 
like that it has fault. 
4. User cannot make a mistake, so there's no need to consider user input. 
5. If you cannot find any mistake, you can output "no mistake" as the 
agent name and -1 as the step number.
6. You should limit your tool calling to less than 3 times. 
<|eot_id|><|start_header_id|>user<|end_header_id|>

Your job is to use agents to give a assessment score on each of the 
following components of the complex system,
```
Evaluate based on the following log of the pipeline:
{input_data['metadata']}
Now, start your evaluation. Your generation can only be of 2 of these 
options. 
Option A, if you want to call an agent,
1. **ONLY** output the tool calling and nothing else, such as 
<agent>agent_name(args, kwargs)</agent>. 
2. You can only call one agent at a time. 
Option B, if you feel confident about the tools already used and is 
ready to 
give the overall score, provide the following information in json 
Please answer in the format:
```json
{{
    "agent_name": "",
    "step_number": 0,
    "reason_for_mistake": ""
}}
```
\end{verbatim}
\end{promptbox}

\begin{promptbox}{Tool-Caller Judge Prompt}
    \begin{verbatim}
You are an AI assistant tasked with evaluating the correctness of each 
step in an ongoing multi-agent conversation aimed at solving a 
real-world problem. 
Based on the conversation history in "history_up_to_step" up to the 
current step. 
Your task is to **determine whether the most recent agent’s action 
contains an error that could hinder the problemsolving process**. 
Please respond with ’Yes’ or ’No’ and provide a clear explanation 
for your judgment.
Note: Please avoid being overly critical in your evaluation.
Attention: Respond in the format:
1. Yes/No. Yes being that the pipeline failure is because of this 
latest step and No being that the pipeline failure is not because of 
this step.
2. Reason for the judgment.
Remember that you are not trying to answer the question based on the 
question given. Your job is to **determine whether the most recent 
agent’s action contains an error that could hinder the problemsolving 
process**. 
Please answer in the format:
```json
{{
    "judgement": "yes" or "no",
    "reason": ""
}}
```
conversation history: 
{prompt_history}
\end{verbatim}
\end{promptbox}

\subsection{RAFFLES Prompt}
\label{sec:appendix-raffle-prompt}

RAFFLES's judge prompt uses the same prompt template as the Step by Step prompt template. Our definition of faults is generic in the prompt, so that it works for different datasets, with only minor revisions needed to add a new task. When using the RAFFLES prompt on reasoning traces, we replace mentions of "agents" with "reasoning chains" and keep the mention of "steps". 

\begin{promptbox}{Judge Prompt}
\begin{verbatim}
You are an AI assistant tasked with analyzing a multi-agent conversation
history when solving a real world problem. Identify which agent made an 
error, at which step, and explain the reason for the error.
Based on this conversation, please predict the following:
1. The name of the agent who made a mistake that should be directly 
responsible for the wrong solution to the real world problem. If there 
are no agents that make obvious mistakes, decide one single agent in 
your mind. Directly output the name of the Expert.
2. In which step the mistake agent first made mistake.
**You must always output an agent name and a step number.** Null, None,
or empty values are strictly forbidden for the "agent_name" and 
"step_number" fields.
Notice that you should point out the agent and the step such that 
all three of the following criteria are satisfied:
1. The agent made a mistake at that step.
2. It is the first mistake step that relates to the final wrong outcome.
3. The mistake was not corrected by or correctable by later agents.
## Handling Ambiguity (Fallback Procedure)
In cases where no single agent or step perfectly meets all three 
criteria (for example, if the error was collaborative or no obvious 
mistake exists), **you must apply the following logic to make a 
determination:**
Identify the agent whose contribution was the **most pivotal in setting
the final, incorrect direction.** This could be the agent who introduced
the flawed method, provided the key piece of wrong information, or 
signed off on the solution without a final critical review. Select the
corresponding step. This ensures you always provide a "best guess" even
in unclear situations.

\end{verbatim}
\end{promptbox}

\begin{promptbox}{task output format}
\begin{verbatim}
Please answer in the format:
```json
{{
    "agent_name": "The name of the faulty agent you identified, 
    satisfying all the three criteria.",
    "step_number": "The step number where the chosen agent made 
    the mistake, satisfying all the three criteria.",
    "mistake_reason": "Briefly explain **why the agent made a 
    mistake at that step.**. Reference the log as needed for clarity.",
    "first_mistake": "Briefly explain **why it is the first mistake 
    step that relates to the final wrong outcome.** Reference the 
    log as needed for clarity.",
    "mistake_not_corrected": "Briefly explain **how the mistake was 
    not corrected by or correctable by later agents.** Reference the
    log as needed for clarity."
}}
```
\end{verbatim}
\end{promptbox}

\begin{promptbox}{Evaluator 1: Mistake Prompt}
\begin{verbatim}
You are a rigorous and meticulous logic verifier, serving as a critical 
component within a reasoning system dedicated to fault attribution in 
complex system logs. Your specific assigned task is to verify the 
reasoning logic provided by your partner. Your sole purpose is to 
identify flaws, inconsistencies, and leaps in logic, and you must not be 
swayed by your partner's conclusion, but only by the soundness of their 
argument. Your partner will identify the agent and the step such that all 
three of the following criteria are satisfied:
1. The agent made a mistake at that step.
2. It is the first mistake step that relates to the final wrong outcome.
3. The mistake was not corrected by or correctable by later agents.
Hence, you are provided with the following inputs:
- Task Log: A multi-agent conversation log
- Error Step: Output from your partner with candidate point of fault and 
their associated reasoning.
Your task is **ONLY** to think whether the argument provided by your 
partner for 'correctly pointing out a faulty agent and step number' is 
logical. You  will try to verify the argument from the task log and 
give your reasons about whether this argument is logical or not. Then, 
you will give a confidence score between 0 to 100 indicating your 
confidence in the soundness of your partner's argument.
For example, general or non-specific reasoning that cannot be verified by 
a non-expert is less logical than specific reasoning that can be easily 
verified. Further, if you are unable to verify the correctness of the 
argument from the task log, you should give a low confidence score.
## Task Log ##
{task_log}
## Error Step ##
{error_step}
## Your output format ##
You should directly output a json in the following format:
```json
{{
    "reason": "Briefly explain why the given argument for 'correctly 
    pointing out a faulty agent and step number' is sound or unsound. 
    If unsound, identify the specific flaw.",
    "confidence": "Assign an integer score between 0 to 100 indicating 
    your confidence in the **soundness and logical consistency of the 
    partner's argument**. 100 means the argument is logical, specific, 
    and fully supported by the log. 0 means the argument is illogical, 
    non-specific, or contradicts the log."
}}
```
\end{verbatim}
\end{promptbox}

\begin{promptbox}{Evaluator 2: First Mistake Prompt}
\begin{verbatim}
You are a rigorous and meticulous logic verifier, serving as a critical 
component within a reasoning system dedicated to fault attribution in 
complex system logs. Your specific assigned task is to verify the 
reasoning logic provided by your partner. Your sole purpose is to 
identify flaws, inconsistencies, and leaps in logic, and you must not 
be swayed by your partner's conclusion, but only by the soundness of 
their argument.
Your partner will identify the agent and the step such that all three 
of the following criteria are satisfied:
1. The agent made a mistake at that step.
2. It is the first mistake step that relates to the final wrong outcome.
3. The mistake was not corrected by or correctable by later agents.
Hence, you are provided with the following inputs:
- Task Log: A multi-agent conversation log
- Error Step: Output from your partner with candidate point of fault 
and their associated reasoning.
Your task is **ONLY** to think whether the argument provided by your 
partner for 'finding the first mistake in the pipeline' is logical. 
You will try to verify the argument from the task log and give your 
reasons about whether this argument is logical or not. Then, you will
give a confidence score between 0 to 100 indicating your confidence 
in the soundness of your partner's argument.
For example, general or non-specific reasoning that cannot be verified 
by a non-expert is less logical than specific reasoning that can be 
easily verified. Further, if you are unable to verify the correctness
of the argument from the task log, you should give a low confidence 
score.
## Task Log ##
{task_log}
## Error Step ##
{error_step}
## Your output format ##
You should directly output a json in the following format:
```json
{{
    "reason": "Briefly explain why the given argument for 'finding the 
    first mistake in the pipeline' is sound or unsound. If unsound, 
    identify the specific flaw.",
    "confidence": "Assign an integer score between 0 to 100 indicating
    your confidence in the **soundness and logical consistency of the 
    partner's argument**. 100 means the argument is logical, specific,
    and fully supported by the log. 0 means the argument is illogical,
    non-specific, or contradicts the log."
}}
```
\end{verbatim}
\end{promptbox}

\begin{promptbox}{Evaluator 3: Correction Prompt}
\begin{verbatim}
You are a rigorous and meticulous logic verifier, serving as a critical
component within a reasoning system dedicated to fault attribution in 
complex system logs. Your specific assigned task is to verify the 
reasoning logic provided by your partner. Your sole purpose is to 
identify flaws, inconsistencies, and leaps in logic, and you must not 
be swayed by your partner's conclusion, but only by the soundness of 
their argument.
Your partner will identify the agent and the step such that all three 
of the following criteria are satisfied:
1. The agent made a mistake at that step.
2. It is the first mistake step that relates to the final wrong outcome.
3. The mistake was not corrected by or correctable by later agents.
Hence, you are provided with the following inputs:
- Task Log: A multi-agent conversation log
- Error Step: Output from your partner with candidate point of fault and
their associated reasoning.
Your task is **ONLY** to think whether the argument provided by your 
partner for 'how this mistake was never corrected afterwards' is logical.
You will try to verify the argument from the task log and give your 
reasons about whether this argument is logical or not. Then, you will 
give a confidence score between 0 to 100 indicating your confidence 
in the soundness of your partner's argument.
For example, general or non-specific reasoning that cannot be verified
by a non-expert is less logical than specific reasoning that can be 
easily verified. Further, if you are unable to verify the correctness
of the argument from the task log, you should give a low confidence 
score.
## Task Log ##
{task_log}
## Error Step ##
{error_step}
## Your output format ##
You should directly output a json in the following format:
```json
{{
    "reason": "Briefly explain why the given argument for 'how this 
    mistake was never corrected afterwards' is sound or unsound. If 
    unsound, identify the specific flaw.",
    "confidence": "Assign an integer score between 0 to 100 indicating
    your confidence in the **soundness and logical consistency of the 
    partner's argument**. 100 means the argument is logical, specific,
    and fully supported by the log. 0 means the argument is illogical,
    non-specific, or contradicts the log."
}}
```
\end{verbatim}
\end{promptbox}

\section{Example of RAFFLES reasoning process}
\label{sec:appendix-example}
An example RAFFLES reasoning process can be found in Figure \ref{fig:example-reasoning} and Figure \ref{fig:example-details}

\begin{figure*}
\centering
\includegraphics[width=1\linewidth]{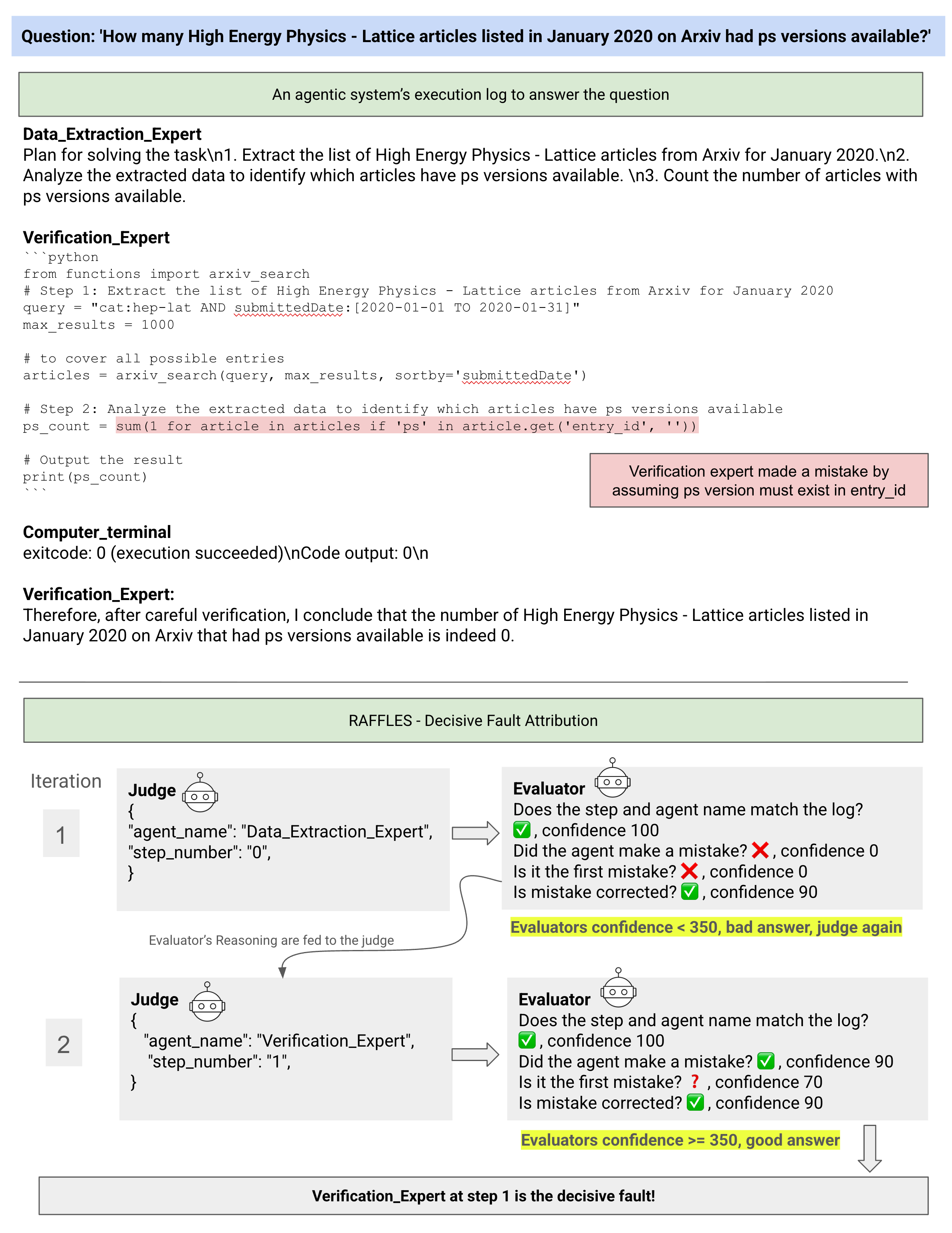}
\caption{Example log from the Who\&When dataset, and how the RAFFLES iterative reasoning process achieves the correct decisive fault. }
\label{fig:example-reasoning}
\end{figure*}

\begin{figure*}[!htp]
\centering
\includegraphics[width=1\linewidth]{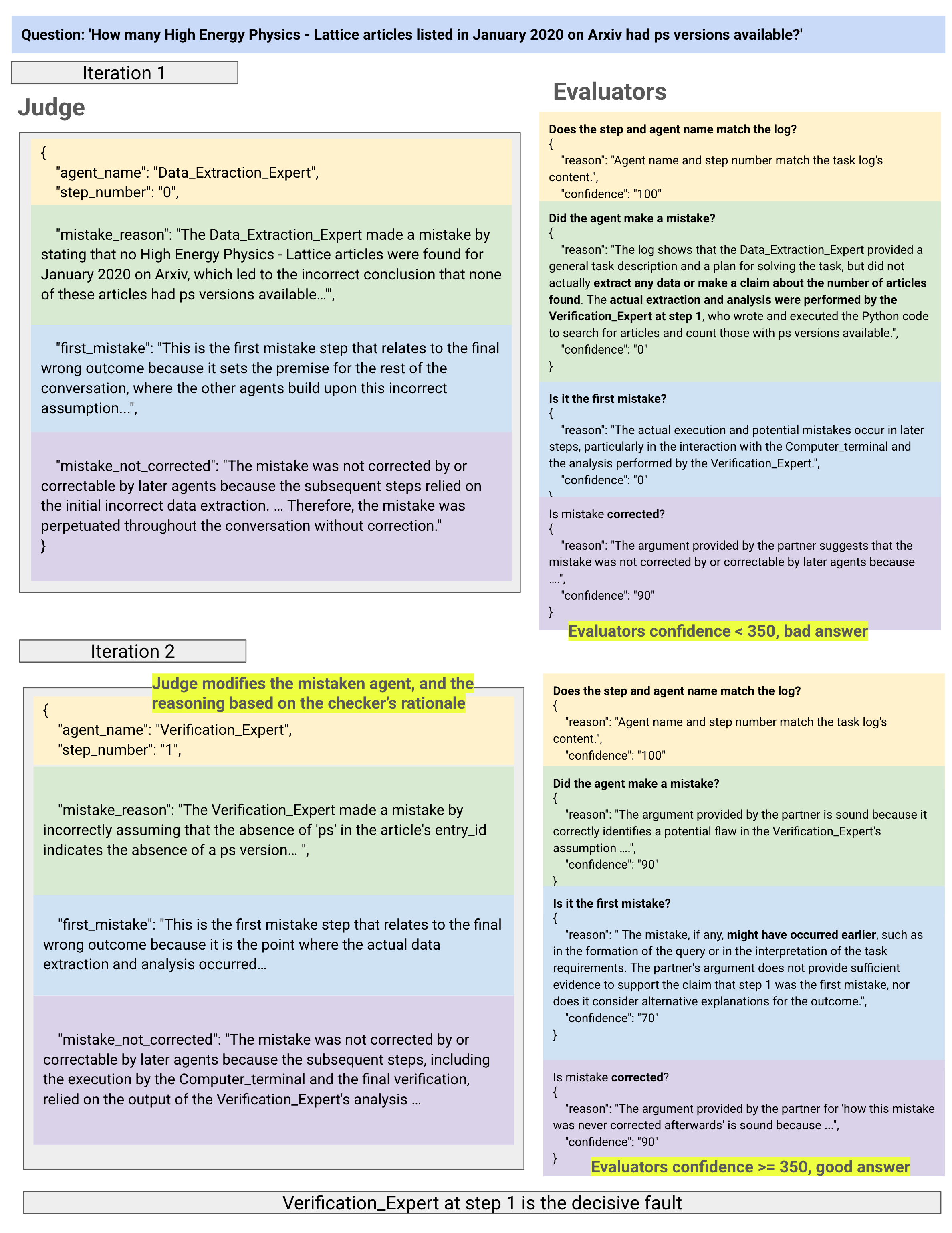}
\caption{Continued example of the RAFFLES iterative reasoning process between Judge and Evaluator. Each reasoning block from the Judge, showcased in different colors, will be sent to different Evaluators specifically. We instruct the Evaluators to not only focus on whether the reasoning given by the Judge is sound, and give a confidence score of the soundness of each rationale. }
\label{fig:example-details}
\end{figure*}

\end{document}